\documentclass[journal]{IEEEtran}
\usepackage{amsmath,amssymb,amsfonts}
\usepackage{cases}
\usepackage{color,xcolor}
\usepackage{graphicx}
\usepackage{subfigure}
\usepackage{algorithm}
\usepackage{algorithmic}
\usepackage{epstopdf}
\usepackage{multirow}
\usepackage{rotating}
\usepackage{booktabs}
\usepackage{tikz}
\usepackage{newtxmath}
\usepackage{acronym}
\usepackage{epstopdf}
\usepackage{xurl}
\usepackage{pifont} 
\usepackage[demo,
            export]{adjustbox}  
\usepackage{tabularray}
\usepackage[table]{xcolor}
\usepackage{colortbl}
\usepackage[colorlinks,
linkcolor=blue,
anchorcolor=blue,
citecolor=blue
]{hyperref}
\usepackage{amssymb}
\usepackage[colorlinks,linkcolor=blue]{hyperref}
\usepackage{array} 
\usepackage{catchfilebetweentags}
\definecolor{tableTitleColor}{rgb}{0.8,0.8,0.8}
\definecolor{SEBEM}{RGB}{183, 222, 232}
\definecolor{SETEM}{RGB}{215, 253, 225}
\definecolor{SENRM}{RGB}{255,242,204}
\definecolor{SEIRM}{RGB}{196,196,196}
\newtheorem{Theorem}{Theorem}[section]

\newtheorem{Remark}[Theorem]{Remark}

\begin{document} 

\title{Lightweight Deep Unfolding Networks with Enhanced Robustness for Infrared \\Small Target Detection}
	
\author{Jingjing Liu, \IEEEmembership{Member,~IEEE}, Yinchao Han, Xianchao Xiu, \IEEEmembership{Member,~IEEE}, Jianhua Zhang, \\ \IEEEmembership{Senior Member,~IEEE}, and Wanquan Liu, \IEEEmembership{Senior Member,~IEEE}

\thanks{This work was supported in part by the National Natural Science Foundation of China under Grant 62204044 and Grant 12371306, and in part by the State Key Laboratory of Integrated Chips and Systems under Grant SKLICS-K202302. (\textit{Corresponding author: Xianchao Xiu}.)}
\thanks{J. Liu, Y. Han, and J. Zhang are with the Shanghai Key Laboratory of Automobile Intelligent Network Interaction Chip and System,  School of Microelectronics,  Shanghai University, Shanghai 200444,  China (e-mail: \{jjliu, ychan, jhzhang\}@shu.edu.cn).}
\thanks{X. Xiu is with the School of Mechatronic Engineering and Automation,  Shanghai University,  Shanghai 200444, China (e-mail: xcxiu@shu.edu.cn).}
\thanks{W. Liu is with the School of Intelligent Systems Engineering, Sun Yat-sen University, Guangzhou 510275, China (e-mail: liuwq63@mail.sysu.edu.cn).}
}

\maketitle	
\begin{abstract}
Infrared small target detection (ISTD) is one of the key techniques in image processing. Although deep unfolding networks (DUNs) have demonstrated promising performance in ISTD due to their model interpretability and data adaptability, existing methods still face significant challenges in parameter lightweightness and noise robustness. In this regard, we propose a highly lightweight framework based on robust principal component analysis (RPCA) called L-RPCANet. Technically, a hierarchical bottleneck structure is constructed to reduce and increase the channel dimension in the single-channel input infrared image to achieve channel-wise feature refinement, with bottleneck layers designed in each module to extract features. This reduces the number of channels in feature extraction and improves the lightweightness of network parameters. Furthermore, a noise reduction module is embedded to enhance the robustness against complex noise. In addition, squeeze-and-excitation networks (SENets) are leveraged as a channel attention mechanism to focus on the varying importance of different features across channels, thereby achieving excellent  performance while maintaining both lightweightness and robustness. Extensive experiments on the ISTD datasets validate the superiority of our proposed method compared with state-of-the-art methods covering RPCANet, DRPCANet, and RPCANet++. The code will be available at \url{https://github.com/xianchaoxiu/L-RPCANet}.
\end{abstract}


\IEEEpeerreviewmaketitle

\section{Introduction}
\IEEEPARstart{U}{nlike} traditional visible light imaging techniques, infrared imaging records the thermal radiation naturally emitted by objects and captures ambient information \cite{yue2023dif,li2023lrrnet}. This enables reliable detection of small targets even in scenarios with strong electromagnetic interference and darkness \cite{lin2024learning}. Consequently, infrared small target detection (ISTD) has attracted widespread attention, with applications ranging from remote sensing, intelligent transportation, aerospace, and medicine, see \cite{yuan2024sctransnet,wang2024vivo,liu2025enhancing,li2025ilnet,liu2025graph}.

As a fundamental image processing task, ISTD faces two major challenges \cite{zhao2022single,kumar2025small}. On the one hand, targets in infrared images typically occupy an extremely limited number of pixels and have a relatively low signal-to-noise ratio (SNR), resulting in limited target texture information \cite{gao2018infrared}. On the other hand, the computing resources of mobile devices are generally limited, making it difficult to meet the requirements of real-time detection systems \cite{zhang2025m4net}. 
During the past few decades, numerous ISTD methods have been developed, which can be generally divided into three categories: model-based methods \cite{gao2013infrared}, data-driven methods \cite{wu2023uiu}, and model-data-driven methods \cite{wu2024rpcanet}.

Model-based ISTD methods integrate the detection task into a physical model, where targets are represented as compact heat sources embedded in a background with statistical characteristics \cite{gao2018infrared,liu2025tensor}. Due to its simple mathematical formulation, low-rank representation has been broadly used in computer vision and industrial engineering \cite{sun2023learning,xiu2024efficient}. The core idea of low-rank ISTD methods is to characterize the slowly varying but highly correlated background as a low-rank component, and extremely small-sized but energy-isolated targets as sparse components \cite{zhu2020tnlrs}. Building on this, Gao \textit{et al.} \cite{gao2013infrared} proposed a baseline method called the infrared patch image (IPI), which leverages the classic principal component analysis (PCA) to simultaneously separate the low-rank background matrix and the sparse target matrix, thereby extracting target texture information. However, the IPI converts infrared images into two-dimensional matrices, which destroys the natural spatial-spectral neighborhood correlation within the image. Instead of using matrices to characterize the data, Zhang \textit{et al.} \cite{zhang2019infrared} utilized the low-rank tensor to capture the background and detect targets through a nonconvex method based on the partial sum of the tensor nuclear norm (PSTNN). By applying non-local self-similarities, the above two methods can simultaneously achieve background estimation and target extraction while suppressing large-scale textures and fixed pattern noise \cite{lin2024learning}. Different from low-rank methods, Wei \textit{et al.} \cite{wei2016multiscale} suggested using neighborhood windows of varying sizes to scan pixel by pixel to form a multilayer saliency map, followed by adaptive threshold segmentation to extract infrared small targets. This method, called multiscale patch-based contrast measurement (MPCM), also performs well numerically. Although model-based ISTD methods offer high interpretability, they are often affected by noise and complex background, and involve a large number of hyperparameters when modeling, resulting in poor robustness and generalization  \cite{liu2023combining,pang2024lrta,luo2024revisiting}.

Data-driven ISTD methods achieve promising pixel-level detection accuracy through end-to-end learning with a variety of semantic segmentation networks \cite{chen2022local,zhang2024irprunedet}. For example,
Zhang \textit{et al.} \cite{zhang2023attention} integrated dilated spatial pyramid pooling to ResNet-101 \cite{he2016deep}, then fused low-level details and high-level semantics through convolution and up-sampling. This method is called attention-guided pyramid context networks (AGPCNet), which significantly improves the performance of ISTD compared to IPI and PSTNN. Later, Liu \textit{et al.} \cite{liu2024infrared} proposed a simple multiscale head to the plain U-Net \cite{ronneberger2015u} named MSHNet by considering ResNet-50 as the encoder. At the stem layer, the first convolution is replaced with a rotation-equivariant windmill convolution, thus achieving lightweight multiscale semantic segmentation. In addition, Wu \textit{et al.} \cite{wu2023uiu} employed a nested structure of U-Net in U-Net to enclose an isomorphic sub-network within the encoding-decoding process, which is abbreviated as UIUNet. This design allows the outer network to preserve global contours, while the inner network focuses on pixel hotspots. However, there are still some practical challenges, such as the computationally intensive fusion phase and the large amount of data required for self-training \cite{yang2025deep}.

Compared to model-based and data-driven ISTD methods, model-data-driven ISTD methods have the ability to take into account their advantages \cite{wu2024extrapolated,deng2025deepsn}. The design of deep unfolding networks (DUNs) to handle ISTD is a popular direction that bridges the gap between iterative algorithms and neural networks, leveraging mechanistic modeling and network-based solutions \cite{shlezinger2023model,joukovsky2023interpretable,xiong2025drpca}.
Recently, Wu \textit{et al.} \cite{wu2024rpcanet} treated the ISTD task as a robust PCA (RPCA) problem \cite{candes2011robust}, and extended the optimization steps to DUNs. This method is dubbed RPCANet, and enjoys good interpretability. Xiong \textit{et al.} \cite{xiong2025drpca} proposed a dynamic RPCA network (DRPCANet), which employs a dynamic parameter generation mechanism and a dynamic residual group module to separate sparse targets from the low-rank background to achieve efficient detection of infrared small targets. To further improve detection efficiency and convergence speed, Wu \textit{et al.} \cite{wu2025rpcanet++} proposed RPCANet++ based on RPCANet, which incorporates a memory-augmented module to preserve background features during background extraction and a deep contrastive prior module to accelerate target extraction during target approximation. In summary, these studies highlight the effectiveness of DUNs in balancing performance, interpretability, and input adaptability, which motivates our interest in applying the evolving RPCA framework for the ISTD task.

\begin{figure}[t]
    \centering
    \includegraphics[width = 0.7\columnwidth]{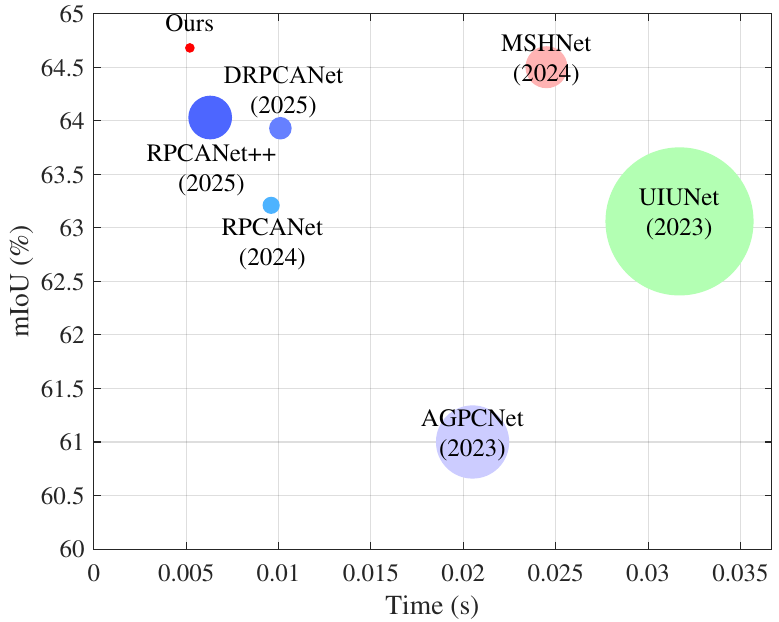}
    \caption{Comparison of detection performance (mIoU), and inference time (s) on GPU of various methods on the IRSTD-1k dataset. The size of the bubbles represents the number of model parameters.} 
    \label{fig:Time-mIoU} 
\end{figure}

Motivated by these observations, we propose a lightweight DUNs-based method with enhanced robustness, abbreviated as L-RPCANet. Compared to the existing RPCANet, DRPCANet, and RPCANet++, our method not only improves the detection performance and noise robustness on ISTD, but also requires fewer parameters, achieves faster GPU inference time, and is more applicable to real-time detection, as shown in Fig. \ref{fig:Time-mIoU}. In general, the main contributions are in three aspects.
\begin{itemize}
    \item We construct a hierarchical bottleneck structure for the dimension decrease and increase in the single-channel input infrared image to achieve the refinement of channel features and design bottleneck layers in the structure to extract the features, addressing the difficulty of extracting small targets. There are fewer channels and parameters in the bottleneck layers, but better in the computational performance of the networks.
    \item We introduce squeeze-and-excitation networks (SENets) as the channel attention mechanism in each module to focus on varying important channel features of the targets. This helps to improve the model's detection performance while maintaining a lightweight architecture and addressing the low contrast of small targets in infrared images.
    \item We integrate a noise reduction module with SENets to deal with complex noise in infrared images and improve robustness, addressing the issue that ISTD is prone to being disturbed by noise.
\end{itemize}

The rest of the paper is listed as follows. Section \ref{Related} briefly introduces DUNs and the attention mechanism. Section \ref{Method} presents our model, iterative updates, and detailed networks. Section \ref{Experiment} analyzes the detection results and visualizations. Section \ref{Conclusion} concludes this paper and provides future directions.

\section{Related Work}\label{Related}

\subsection{Deep Unfolding Networks}

Deep unfolding networks (DUNs) inherit physical interpretability and high efficiency by transforming iterative algorithms layer by layer into a learnable architecture \cite{elad2023image,chen2024learning}. Notably, DUNs allow for the replacement of manually tuned hyperparameters and regularization priors with trainable modules, enabling end-to-end optimization \cite{kong2021deep,de2024deep,fu2024rotation}.

In fact, DUNs can be traced back to Gregor and LeCun \cite{gregor2010learning},  who extended the iterative soft thresholding algorithm (ISTA) to learned ISTA (LISTA) by applying feedforward neural networks for compressed sensing (CS) problems. Yang \textit{et al.} \cite{yang2020admm} unfolded the alternating direction method of multipliers (ADMM) algorithm into a learnable framework, called ADMM-CSNet, and achieved favorable reconstruction accuracy. Recently, Sun \textit{et al.} \cite{zhang2018ista} introduced innovative convolutional neural networks (CNNs) to improve the performance of LISTA and the proposed ISTA-Net. Sun \textit{et al.} \cite{you2021ista} constructed an improved version called ISTA-Net++, which enhances feature extraction capabilities by leveraging residual connections and a deeper network architecture. Furthermore, Han \textit{et al.} \cite{han2025dista} proposed the dynamic iterative shrinkage thresholding network, named DISTANet, which transforms the traditional sparse reconstruction method into a dynamic deep learning framework by dynamically generating convolutional weights and threshold parameters.

For the ISTD task, the integration of DUNs with RPCA has been verified to be very promising. However, these methods still have some shortcomings. For example, RPCANet \cite{wu2024rpcanet} lacks channel-wise feature prioritization and noise handling, limiting performance in complex scenarios. DRPCANet \cite{xiong2025drpca} enhances the detection performance by applying the dynamic parameter generation mechanism. However, it is overly dependent on the input features, which leads to false positive targets being detected in some real scenarios. RPCANet++ \cite{wu2025rpcanet++} adopts a fully convolutional architecture, resulting in a slower inference speed than lightweight dense networks. Therefore, our proposed method embeds bottleneck layers and learnable noise reduction modules in the lightweight expansion process to overcome feature blindness and sensitivity to noise of RPCANet, the need to specific input of DRPCANet, as well as the computational cost of RPCANet++.

\subsection{Attention Mechanism}
The attention mechanism \cite{vaswani2017attention} allows neural networks to pay more attention to relevant input regions when generating predictions and is widely used in image classification \cite{guo2023beyond}, semantic segmentation \cite{huang2019ccnet}, and target detection \cite{woo2018cbam}. For ISTD, targets are often missing in deep semantic feature maps due to the limited number of pixels. Therefore, most data-driven methods utilize the attention mechanism to enhance the representation, leading to better performance \cite{yang2025mtmlnet,zhang2024irsam,zhang2025saist}.

For example, UIUNet \cite{wu2023uiu} applies the spatial attention mechanism to generate a pixel-level weight map that accentuates the target region and suppresses background textures. DRPCANet \cite{xiong2025drpca} designs a dynamic residual group module to combine residual learning and a dynamic spatial attention mechanism. This enables the model to better capture contextual changes in the background, thereby achieving more accurate low-rank estimation and small target separation. In addition, RPCANet++ \cite{wu2025rpcanet++} employs a temporal evolution attention mechanism, dynamically assigning weights for cross-stage background features through the gating mechanism. This implicit spatio-temporal attention not only avoids the computational load of explicitly calculating the similarity matrix, but also adaptively focuses on the most valuable channels and spatial position information for low-rank background reconstruction during the iterative expansion process.

It concludes that each attention mechanism has its own characteristics, and they all face major challenges, including high computational complexity, feature selection bias, and training difficulties. To address these issues, our proposed method  integrates lightweight SENets into each submodule, recalibrating channel weights with only global pooling and two fully connected layers, adding almost zero latency while continuously suppressing background textures, and amplifying faint hot spots for cascaded recall gains.

\section{The Proposed Method}\label{Method}

\begin{figure*}[t]
    \centering
    \includegraphics[width = 0.95\textwidth]{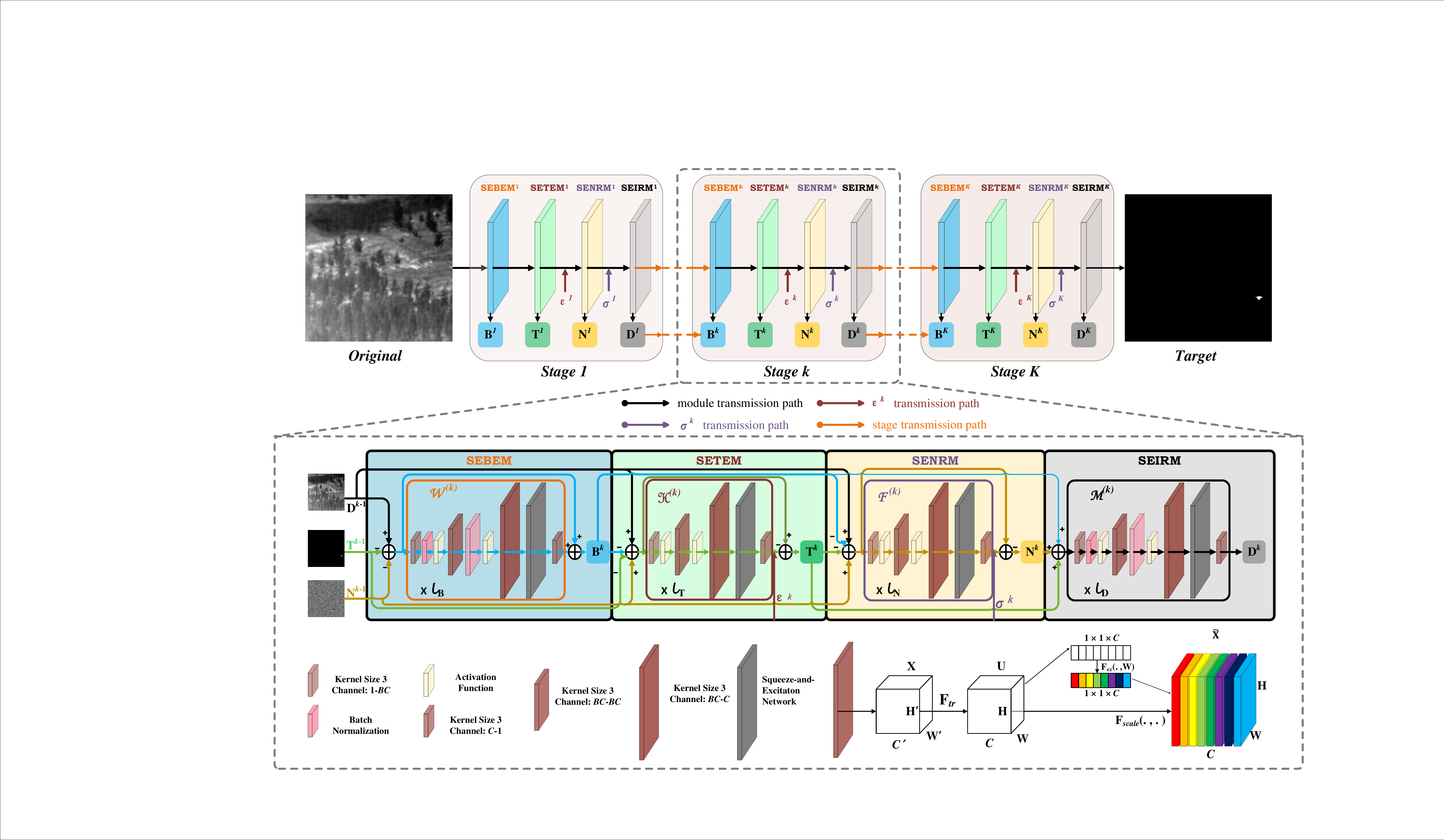} 
    \caption{Overall structure of our proposed L-RPCANet, which is composed of K stages. The detail network structure of the single stage concludes background estimation module with SENets (\texttt{SEBEM}), targets extraction module with SENets (\texttt{SETEM}), noise reduction module with SENets (\texttt{SENRM}), and image reconstruction module with SENets (\texttt{SEIRM}). The network in the lower-right corner shows the structure of SENets.} 
    \label{fig:framework} 
\end{figure*}

\subsection{Problem Formulation} \label{Problem Formulation}

Given an infrared image $\mathbf{D}$, according to the physical prior, it can be decomposed into the background part $\mathbf{B}$, the target part $\mathbf{T}$, and the noise part $\mathbf{N}$ as
\begin{equation}\label{Eq.(1)}
\mathbf{D} = \mathbf{B} + \mathbf{T} + \mathbf{N},
\end{equation}
where $\mathbf{D}, \mathbf{B}, \mathbf{T}, \mathbf{N} \in \mathbb{R}^{m \times n}$.
Generally speaking, the background in infrared images has the low-rank property, small targets are sparse, and the noise follows a Gaussian distribution. Based on RPCA \cite{candes2011robust}, the ISTD task is described as the following optimization framework
\begin{equation}\label{Eq.(2)}
\begin{aligned}
\min_{\mathbf{B}, \mathbf{T}, \mathbf{N}} \quad & \|\mathbf{B}\|_* + \lambda \|\mathbf{T}\|_1 + \frac{\mu}{2}\|\mathbf{N}\|_\textrm{F}^2  \\
\textrm{s.t.}  ~\quad &  \mathbf{D} = \mathbf{B} + \mathbf{T} + \mathbf{N},
\end{aligned}
\end{equation}
where $\lambda, \mu>0$ are the trade-off parameters, $\|\mathbf{B}\|_*$ is the nuclear norm, \textit{i.e.}, the sum of all singular values, $\|\mathbf{T}\|_1$ is the $\ell_1$ norm, \textit{i.e.}, the sum of all absolute values of all elements, and $\|\mathbf{N}\|_\textrm{F}^2$ is the squared Frobenius norm, \textit{i.e.}, the sum of the squares of all elements.

However, in real-world infrared scenes, the background and targets are often very complex, and thus a single norm cannot capture practical structures. Besides, the noise is not necessarily Gaussian \cite{bouwmans2017decomposition}. In this paper, we intend to adopt more general functions $\mathcal{L}(\mathbf{B})$, $\mathcal{S}(\mathbf{T})$, and $\mathcal{R}(\mathbf{N})$ to describe the background, targets, and noise. Therefore, the above problem (\ref{Eq.(2)}) is rewritten as 
\begin{equation} \label{Eq.(3)}
   \begin{aligned}
         \min\limits_{\mathbf{B}, \mathbf{T}, \mathbf{N}} \quad & \mathcal{L}(\mathbf{B}) + \lambda \mathcal{S}(\mathbf{T}) + \mu \mathcal{R}(\mathbf{N}) \\        
         \textrm{s.t.}~\quad &  \mathbf{D} = \mathbf{B} + \mathbf{T} + \mathbf{N}.
   \end{aligned}
\end{equation}
Unlike RPCANet \cite{wu2024rpcanet}, DRPCANet \cite{xiong2025drpca}, and RPCANet++ \cite{wu2025rpcanet++}, there are two differences.
\begin{itemize}
    \item A noise component $\mathbf{N}$ is enforced into the decomposition constraint, which makes our model more robust.
    \item A general function $\mathcal{R}(\mathbf{N})$ with a parameter $\mu$ is added to the objective, which can be learned via neural networks.
\end{itemize}

From the optimization perspective, the constrained problem (\ref{Eq.(3)}) has the following equivalent unconstrained version
\begin{equation}\label{Eq.(4)}
    \begin{aligned} 
        \mathcal{L}(\mathbf{B}, \mathbf{T}, \mathbf{N}) & = \mathcal{L}(\mathbf{B}) + \lambda\mathcal{S}(\mathbf{T}) +\mu\mathcal{R}(\mathbf{N}) \\
        & \quad + \frac{\alpha}{2}\| \mathbf{D} - \mathbf{B} - \mathbf{T} - \mathbf{N}\|_\textrm{F}^2,
    \end{aligned}
\end{equation}
where $\alpha>0$ is the penalty parameter.
In the next subsection, we describe how to update $\mathbf{B}$, $\mathbf{T}$, and $\mathbf{N}$ using neural networks. It is worth pointing out that $\mathbf{D}$ needs to be updated through the image reconstruction module, which is often overlooked in traditional iterative methods.

\subsection{Network Architecture}

Before proceeding, let us first introduce the squeeze-and-excitation networks (SENets). The basic structure of the SE building block is provided in the lower right corner of Fig. \ref{fig:framework}.  The squeeze operation compresses the feature maps across the spatial dimensions using global average pooling. This operation generates a channel descriptor for each channel, which captures the global distribution of feature responses across the entire spatial extent of the feature map. The excitation operation uses a self-gating mechanism to learn sample-specific activations for each channel.

\subsubsection{Updating $\mathbf{B}$} 

For the background part $\mathbf{B}$, the subproblem is formulated as 
\begin{equation} \label{Eq.(5)}
    \mathbf{B}^k = \arg\min_{\mathbf{B}} ~ \mathcal{L}(\mathbf{B}) + \frac{\alpha}{2}\| \mathbf{D}^{k-1} - \mathbf{B} - \mathbf{T}^{k-1}- \mathbf{N}^{k-1}\|_\textrm{F}^2.
\end{equation}

Assuming $\mathcal{L}(\mathbf{B})=\|\mathbf{B}\|_*$, the singular value threshold (SVT) algorithm \cite{cai2010singular} can be easily used, with its operator denoted as $\mathrm{prox}_{\alpha \|\cdot\|_*}(\cdot)$. However, this is computationally inefficient for large-scale datasets and poses challenges in the selection of $\alpha$. Furthermore, $\mathcal{L}(\mathbf{B})$ may have other low-rank structures, such as the weighted nuclear norm \cite{gu2017weighted} and non-convex overlapped nuclear norm \cite{yao2022low}. To better simulate the low-rank operator including $\mathrm{prox}_{\alpha \|\cdot\|_*}(\cdot)$ and learn tuning parameters, we resort to the residual structure $\mathrm{proxNet}(\cdot)$, which follows a similar idea as \cite{zhang2018ista,wu2024rpcanet}. The solution of problem (\ref{Eq.(5)}) is 
\begin{equation} \label{Eq.(6)}
    \begin{aligned}
        \mathbf{B}^{k} 
        & = \mathrm{proxNet}(\mathbf{D}^{k-1} - \mathbf{T}^{k-1} - \mathbf{N}^{k-1} )\\
        & \approx \mathbf{D}^{k-1} - \mathbf{T}^{k-1} - \mathbf{N}^{k-1} \\
        & \quad + \mathcal{W}^{k}(\mathbf{D}^{k-1} - \mathbf{T}^{k-1} - \mathbf{N}^{k-1}),
    \end{aligned}
\end{equation}
where $\mathcal{W}^{k}(\cdot)$ denotes the $3 \times 3$ convolution group.

As shown in Fig. \ref {fig:framework}, \texttt{SEBEM} is applied to estimate the background $\mathbf{B}^k$. 
Specifically, $\mathcal{W}^{k}(\cdot)$ involves the search for twice channel mapping relationships: from 1 to $BC$ and from $BC$ to $C$. The background features are extracted at the bottleneck layers with $BC$ channels. Here, BN stands for batch normalization and ReLU stands for rectified linear unit \cite{nair2010rectified}. 
After processing by SENets, the feature map with $C$ channels is adjusted to the weights of each channel, thereby enhancing the representation capabilities.

\subsubsection{Updating $\mathbf{T}$} 

For the target part $\mathbf{T}$, the subproblem is written in the form of
\begin{equation}     \label{Eq.(7)}
    \mathbf{T}^k = \arg\min_{\mathbf{T}} ~\lambda\mathcal{S}(\mathbf{T}) + \frac{\alpha}{2}\| \mathbf{D}^{k-1} - \mathbf{B}^k - \mathbf{T}- \mathbf{N}^{k-1}\|_\textrm{F}^2.
\end{equation}      

If $\mathcal{S}(\mathbf{T})=\|\mathbf{T}\|_1$, many efficient optimization algorithms have been developed, such as \cite{beck2009fast,li2018highly}.
Unfortunately, they cannot be generalized to other more complex sparse structures. To obtain a simpler and more intuitive solution, we approximate $\mathcal{S}(\mathbf{T})$ using the Taylor expansion as
\begin{equation}     \label{Eq.(8)}
\begin{aligned}
      \mathbf{T}^k & = \arg\min_{\mathbf{T}} ~\frac{\lambda L_T}{2}\| \mathbf{T} - \mathbf{T}^{k-1} - \frac{1}{L_T}\nabla \mathcal{S}(\mathbf{T}^{k-1})\|_\textrm{F}^2\\
    &\quad \quad + \frac{\alpha}{2}\| \mathbf{D}^{k-1} - \mathbf{B}^k - \mathbf{T}- \mathbf{N}^{k-1}\|_\textrm{F}^2,
\end{aligned}
\end{equation}   
where $\nabla \mathcal{S}(\mathbf{T}^{k-1})$ is the gradient of $\mathbf{T}^{k-1}$, and $L_T$ the Lipschitz constant of $\mathcal{S}(\mathbf{T}^{k-1})$. By taking the derivative and setting it equal to zero, a closed-form solution is derived, given by
\begin{equation}\label{Eq.(9)}
    \begin{aligned}
         \mathbf{T}^k & = \frac{\lambda L_{T}}{\lambda L_{T} + \alpha} \mathbf{T}^{k-1}+ \frac{\alpha}{\lambda L_{T} + \alpha} (\mathbf{D}^{k-1} - \mathbf{B}^k -\mathbf{N}^{k-1}) \\ 
        &\quad- \frac{\lambda}{\lambda L_{T} + \alpha} \nabla \mathcal{S}(\mathbf{T}^{k-1}).
    \end{aligned}
\end{equation}

For ease of description, denote
\begin{equation} 
    \begin{aligned}
\gamma=\frac{\lambda L_{T}}{\lambda L_{T} + \alpha},~\varepsilon=\frac{\lambda}{\lambda L_{T} + \alpha}.
    \end{aligned}
\end{equation}
Then (\ref{Eq.(8)}) is equal to
\begin{equation}     \label{Eq.(11)}
    \begin{aligned}
\mathbf{T}^k & =  \gamma \mathbf{T}^{k-1} + (1 - \gamma)(\mathbf{D}^{k-1} - \mathbf{B}^{k}- \mathbf{N}^{k-1}) \\ 
& \quad - \varepsilon \nabla \mathcal{S}(\mathbf{T}^{k-1}).
    \end{aligned}
\end{equation}
Further, set $\gamma=0.5$ and specify $\varepsilon$ as the learnable parameter $\varepsilon^{k}$ at each stage. Therefore, the solution of problem  (\ref{Eq.(8)}) is 
\begin{equation}     \label{Eq.(12)}
    \begin{aligned}
        \mathbf{T}^{k} & \approx  \mathbf{T}^{k-1} + \mathbf{D}^{k-1} - \mathbf{B}^{k}- \mathbf{N}^{k-1}\\
        &\quad - \varepsilon^{k}\mathcal{H}^{k}(\mathbf{T}^{k-1} + \mathbf{D}^{k-1} - \mathbf{B}^{k}- \mathbf{N}^{k-1}),
    \end{aligned}
\end{equation}
where $\mathcal{H}^{k}(\cdot)$ contains an initial convolution layer to simulate the gradient function $\nabla \mathcal{S}(\cdot)$.

In Fig. \ref{fig:framework}, \texttt{SETEM} is designed to capture the sparse targets $\mathbf{T}^k$. For the gradient function $\nabla \mathcal{S}(\cdot)$, \texttt{SETEM} employs a twice channel mapping and feature extraction in the bottleneck layers similar to \texttt{SEBEM}. However, because the scaling and bias factors in the BN layer are continuously updated during training, which would cause the output-to-input transformation to violate Lipschitz continuity \cite{virmaux2018lipschitz}, BN is not used here. 
In addition, \texttt{SETEM} applies laconic convolution layers that capture the spatial changes of the gradient by learning local features and parameter updates, and ReLU that enhances the model's learning ability of gradient information through nonlinear mapping and effective gradient transfer to simulate the function $\nabla \mathcal{S}(\cdot)$ without BN in the convolution layers.

\subsubsection{Updating $\mathbf{N}$} 
For the noise part $\mathbf{N}$, a closed-form solution can be obtained based on the approximation technique similar to (\ref{Eq.(7)})-(\ref{Eq.(8)}) as follows
\begin{equation}        \label{Eq.(13)}
    \begin{aligned}
        \mathbf{N}^k & = \frac{\mu L_N}{\mu L_N + \alpha} \mathbf{N}^{k-1}  + \frac{\alpha}{\mu L_N + \alpha} (\mathbf{D}^{k-1} - \mathbf{B}^k - \mathbf{T}^k ) \\
        &\quad- \frac{\mu}{\mu L_{N} + \alpha} \nabla \mathcal{G}(\mathbf{N}^{k-1}),
    \end{aligned}
\end{equation}
where $\nabla \mathcal{G}(\mathbf{N}^{k-1})$ and $L_N$ are the corresponding gradient and the Lipschitz constant, respectively.

Let
\begin{equation} 
    \begin{aligned}
\delta=\frac{\mu L_{N}}{\mu L_{\mathcal{N}} + \alpha},~\sigma=\frac{\mu}{\mu L_{N} + \alpha},
    \end{aligned}
\end{equation}
and set $\delta=0.5$, (\ref{Eq.(13)}) is solved as
\begin{equation} 
    \begin{aligned}
        \mathbf{N}^{k} & = \delta \mathbf{N}^{k-1} + (1 - \delta)(\mathbf{D}^{k-1} - \mathbf{B}^k- \mathbf{T}^{k})\\
                &\quad- \sigma \nabla \mathcal{G}(\mathbf{N}^{k-1})\\
        & \approx  \mathbf{N}^{k-1} + \mathbf{D}^{k-1} - \mathbf{B}^{k}- \mathbf{T}^{k}\\
        &\quad - \sigma^{k}\mathcal{F}^{k}(\mathbf{N}^{k-1} + \mathbf{D}^{k-1} - \mathbf{B}^{k}- \mathbf{T}^{k}).
    \label{Eq.(15)}
    \end{aligned}
\end{equation}

 In Fig. \ref{fig:framework} it is found that \texttt{SENRM} takes as input the updated $\mathbf{B}^{k}$ from the last module \texttt{SEBEM}, the updated $\mathbf{T}^{k}$ from the last module \texttt{SETEM}, the noise $\mathbf{N}^{k-1}$, and the reconstruction result $\mathbf{D}^{k-1}$. Although the residual connections and the gradient learning network settings for $\mathcal{F}^{k}(\cdot)$ are similar to those in $\mathcal{H}^{k}(\cdot)$, their different parameters lead to different learning information and structure, thus achieving different extracted targets and noise.

\subsubsection{Updating $\mathbf{D}$} 
For the reconstruction part $\mathbf{D}$, it is updated according to
\begin{equation}     \label{Eq.(16)}
    \begin{aligned}
    \mathbf{D}^k & = \mathbf{B}^{k} + \mathbf{T}^{k}+ \mathbf{N}^{k}\\
     & \approx \mathcal{M}^{k}(\mathbf{B}^{k} + \mathbf{T}^{k} + \mathbf{N}^{k}),
 \end{aligned}
\end{equation}
where $\mathcal{M}^{k}(\cdot)$ uses a simple CNN architecture \cite{zhang2018ffdnet}, which has the same convolution layers but with $l_D= 3$ intermediate layers, as shown in Fig. \ref{fig:framework}. In fact, \texttt{SEIRM} is designed to map the background, targets, and noise into a restored image with a neural network $\mathcal{M}^{k}(\cdot)$.

\begin{Remark}
Inspired by DUNs, the proposed framework can learn the parameters of the convolution kernel to extract features that distinguish low-rank background, sparse targets, and noise, the threshold coefficients to optimize signal separation, and the adaptability parameters of the network structure to improve data adaptability. 
\end{Remark}

\begin{table*}[t]
    \centering
    \setlength{\tabcolsep}{6pt} 
    \renewcommand{\arraystretch}{1.2} 
    \caption{Performance comparisons of different methods on three datasets in terms of mIoU (\%), $\mathrm{F_1} (\%)$, $\mathrm{P_d}$ (\%), and $\mathrm{F_a}$ ($10^{-5}$). \\The best results are marked in \textbf{\textcolor{red}{red}}.}
    \label{performance_comparison}
    \resizebox{\textwidth}{!}
    {
    \begin{tabular}{|l!{\vline width 0.4pt}c!{\vline width 0.4pt}c c c c!{\vline width 0.4pt}c c c c!{\vline width 0.4pt}c c c c!{\vline width 0.4pt}c|}
        \hline
        \multirow{2}{*}{Methods} & \multirow{2}{*}{Params} & \multicolumn{4}{c!{\vline width 0.4pt}}{NUDT-SIRST} & \multicolumn{4}{c!{\vline width 0.4pt}}{SIRST-Aug} & \multicolumn{4}{c!{\vline width 0.4pt}}{IRSTD-1k} & Time (s) \\
        \cline{3-14} 
        & & mIoU $\uparrow$ & $\mathrm{F_1}$ $\uparrow$ & $\mathrm{P_d}$ $\uparrow$ & $\mathrm{F_a}$ $\downarrow$ & mIoU $\uparrow$ & $\mathrm{F_1}$ $\uparrow$ & $\mathrm{P_d}$ $\uparrow$ & $\mathrm{F_a}$ $\downarrow$ & mIoU $\uparrow$ & $\mathrm{F_1}$ $\uparrow$ & $\mathrm{P_d}$ $\uparrow$ & $\mathrm{F_a}$ $\downarrow$ & CPU/GPU \\
        \hline   \hline
        \rowcolor{gray!10} IPI & \multicolumn{1}{c!{\vline width 0.4pt}}{--}  & 34.83  & 51.49 & 92.58 & 7.14 & 21.90 & 35.97 & 80.36 & \textbf{\textcolor{red}{2.20}} & 18.67 & 31.48 & 78.54 & 11.11 & 3.0972/-  \\
        MPCM & \multicolumn{1}{c!{\vline width 0.4pt}}{--} & 25.96 & 40.78 & 78.59 & 7.91 & 19.49 &33.00 &93.58 &3.04 & 14.81 & 25.93 & 69.03 & 6.51  & 0.0624/-  \\
        \rowcolor{gray!10} PSTNN & \multicolumn{1}{c!{\vline width 0.4pt}}{--} & 25.46 & 40.58 & 78.52 & 7.95 & 19.76 & 33.00 & 93.40 & 3.14 & 14.87 & 25.89 & 68.73 & 6.51 & 0.2249/-  \\  
        \hline
        AGPCNet & 12.360M & 85.31 & 92.45 & 97.90 & 4.77 & 72.36 & 83.83 & 99.03 & 35.56 & 61.00 & 75.75 & 89.35 & 5.34 & -/0.0205 \\
        \rowcolor{gray!10} UIUNet & 50.540M & 88.71 & 94.01 & 91.43 & 1.89 & 71.80 & 83.59 & 98.35 & 28.29 & 63.06 & 77.35 & \textbf{\textcolor{red}{93.60}} & 6.57 & -/0.0317 \\
        MSHNet & 4.065M & 89.99 & 93.57 & 96.07 & 2.63 & 71.64 & 84.16 & 90.78 & 23.09 & 64.50 & 77.55 & 91.68 & 4.46 & -/0.0245  \\   \hline
        \rowcolor{gray!10} RPCANet & 0.680M & 89.31 & 94.35 & 97.14 & 2.87  & 72.54 & 84.08 & 98.21 & 34.14 & 63.21 & 77.45 & 88.31 & 4.39 & -/0.0096  \\
        DRPCANet & 1.169M & \textbf{\textcolor{red}{93.12}} & 96.02 & 98.02 & 1.95  & 73.93 & 85.39 & 98.12 & 30.45 & 63.93 & 78.15 & 92.09 & 4.92 & -/0.0101  \\
        \rowcolor{gray!10} RPCANet++ & 4.396M & 92.46 & 96.05 & 98.05 & \textbf{\textcolor{red}{1.44}}  & 73.14 & 84.39 & 97.36 & 32.48 & 64.03 & 77.26 & 89.35 & \textbf{\textcolor{red}{4.28}} & -/0.0063  \\
        \hline       
        \rowcolor{blue!10} \textbf{Ours} & \textbf{\textcolor{red}{0.216M}} & 92.37 & \textbf{\textcolor{red}{96.54}} & \textbf{\textcolor{red}{98.41}} & 1.79 & \textbf{\textcolor{red}{74.56}} & \textbf{\textcolor{red}{85.43}} & \textbf{\textcolor{red}{99.17}} & 29.78 & \textbf{\textcolor{red}{64.68}} & \textbf{\textcolor{red}{78.55}} & 89.39 & 4.66 & -/0.0052  \\
        \hline
    \end{tabular}
    }
\end{table*}

\begin{table}[t]
    \centering
    \setlength{\tabcolsep}{6pt} 
    \renewcommand{\arraystretch}{1.2} 
    \caption{Area under the receiver operating characteristic (ROC) curve comparisons of different methods on three datasets. \\The best results are marked in \textbf{\textcolor{red}{red}}.}
    \label{AUC_comparison_of_different_methods}
    \begin{tabular}{|l |c |c |c |}
    \hline
        Methods  & ~NUDT-SIRST~ & ~SIRST-Aug~ & ~IRSTD-1k~ \\ 
    \hline   \hline
        \rowcolor{gray!10} IPI & 0.8746  & 0.8344 & 0.7946  \\ 
        MPCM & 0.8645 & 0.8246 & 0.7813 \\ 
        \rowcolor{gray!10} PSTNN & 0.8816 & 0.7955 & 0.7451 \\ 
    \hline
        AGPCNet & 0.9712 & 0.9646 & 0.9215 \\ 
        \rowcolor{gray!10} UIUNet  & 0.9547 & 0.9477 & 0.9177 \\ 
        MSHNet  & 0.9900  & 0.9899 & 0.9485 \\         \hline
        \rowcolor{gray!10} RPCANet & 0.9804 & 0.9879 & 0.9346 \\ 
        DRPCANet & 0.9931 & \textbf{\textcolor{red}{0.9935}} & 0.9616 \\
        \rowcolor{gray!10} RPCANet++
        & 0.9954 & 0.9910 & 0.9556 \\ 
    \hline
        \rowcolor{blue!10} \textbf{Ours} & \textbf{\textcolor{red}{0.9987}} & 0.9913 & \textbf{\textcolor{red}{0.9699}} \\
    \hline
    \end{tabular}
\end{table}

\section{Experiments}\label{Experiment}

In this section, sufficient numerical experiments are conducted to compare the proposed method with some benchmark model-based methods including IPI\footnote{\url{https://github.com/gaocq/IPI-for-small-target-detection}} (2013),  MPCM\footnote{\url{https://github.com/wzy-99/MPCM}} (2016), PSTNN\footnote{\url{https://github.com/Lanneeee/Infrared-Small-Target-Detection-based-on-PSTNN}} (2019), data-driven methods including  AGPCNet\footnote{\url{https://github.com/Tianfang-Zhang/AGPCNet}} (2023), UIUNet\footnote{\url{https://github.com/danfenghong/IEEE_TIP_UIU-Net}} (2023), MSHNet\footnote{\url{https://github.com/Lliu666/MSHNet}} (2024), and model-data-driven methods including RPCANet\footnote{\url{https://github.com/fengyiwu98/RPCANet}} (2024), DRPCANet\footnote{\url{https://github.com/GrokCV/DRPCA-Net}} (2025), RPCANet++\footnote{\url{https://github.com/fengyiwu98/RPCANet}} (2025).

Subsection \ref{sec-a} introduces experimental setups. Subsection \ref{sec-b} gives numerical results and detailed analysis. Subsection \ref{sec-c} provides  ablation studies. Subsection \ref{sec-d} presents more discussions.

\subsection{Experimental Setups}\label{sec-a}

\subsubsection{Dataset Description}

The selected datasets include the small datasets, \textit{i.e.}, NUDT-SIRST\footnote{\url{https://github.com/YeRen123455/Infrared-Small-Target-Detection}} and IRSTD-1k\footnote{\url{https://github.com/RuiZhang97/ISNet}}, and the large dataset, \textit{i.e.}, SIRST-Aug\footnote{\url{https://github.com/Tianfang-Zhang/AGPCNet}}. 
These datasets represent a variety of real-world and synthetic infrared imaging scenes, encompassing ignificant differences in target size, intensity and shape, as well as variations in background complexity (such as urban, marine, aerial, and natural landscapes) and sensor characteristics. In addition, NUDT-SIRST and SIRST-Aug images have 256 $\times$ 256 pixels, while IRSTD-1k images have 512 $\times$ 512 pixels.

\subsubsection{Implementation Details} 
Our proposed method is trained for 400 epochs in the PyTorch framework on each dataset using a Nvidia GeForce 4090 GPU. The Adam optimizer is used with an initial learning rate of $10^{-4}$ and a batch size of 8. For the compared methods, directly choose the default parameters according to their codes.

\subsubsection{Loss Function} 
The ISTD task can be divided into target segmentation and infrared image reconstruction, thus the loss function consists of the following two parts: $L_{\text{segmentation}}$ and $L_{\text{fidelity}}$. The former uses SoftIoU \cite{rahman2016optimizing} to evaluate the target segmentation performance, while the latter uses the minimum variance between the initial image and the reconstructed image to evaluate the reconstruction performance. Specifically, the loss function is defined as
\begin{equation}
     \begin{aligned}
         L_{\text{total}} & = L_{\text{segmentation}} + \eta \cdot L_{\text{fidelity}} \\
         & = (1 - \frac{1}{M_t} \sum_{i=1}^{M_t} \frac{\textrm{TP}}{\textrm{FP + TP + FN}}) \\
         &\quad +  \eta \cdot \frac{1}{M_t M} \sum_{i=1}^{M_t} \| \mathbf{D}^K - \mathbf{D} \|_\textrm{F}^2,
     \end{aligned}
\label{Eq.(17)}
\end{equation}
where $i$ and $M_t$ are the number of iterations of the sample, the total training number, $M$ denotes the total pixels per image. 
TP, FP, and FN denote true positive, false positive, and false negative pixel numbers, respectively. Here, $\eta$ balances the contributions of the two terms, which is empirically evaluated in Subsection \ref{sec-d}

\begin{figure*}[t]
    \centering
    \includegraphics[width=\textwidth]{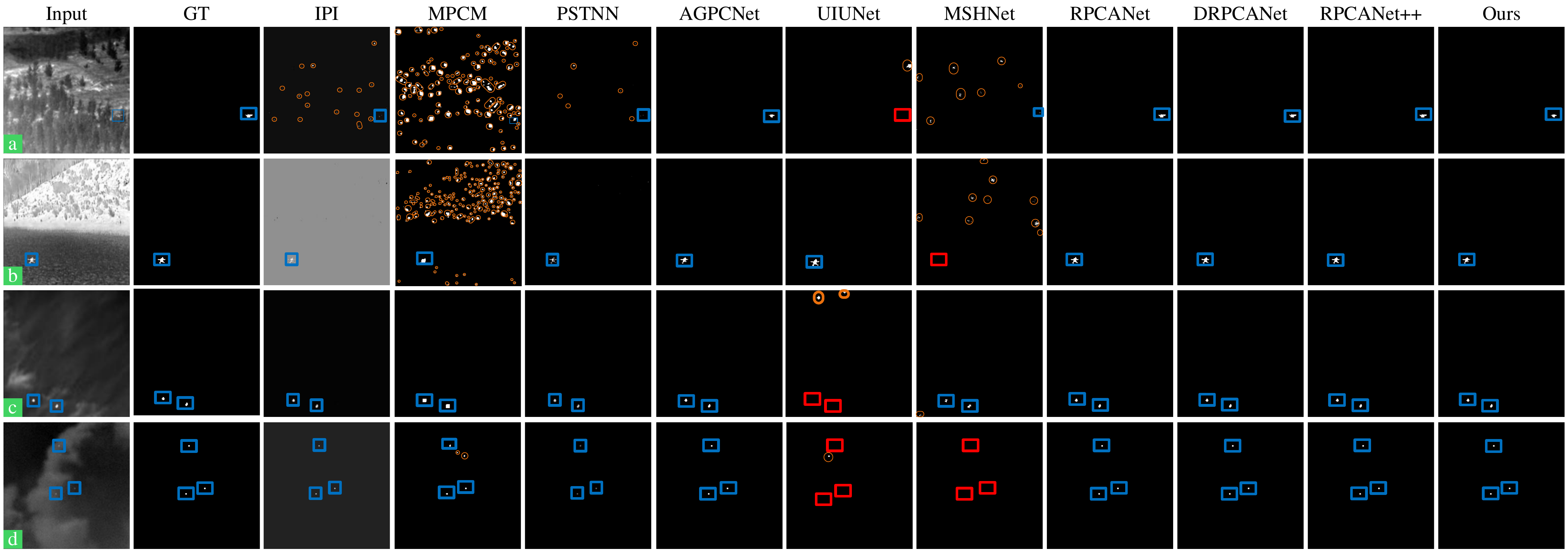}
    \caption{Representative visual results on NUDT-SIRST. There are only one target in (a)-(b), two targets in (c), and three targets in (d). The targets are represented by blue, yellow, and red, indicating true positive targets, false positive targets, and false negative targets, respectively.}
    \label{Representative_visual_results_of_synthetic_dataset_NUDT-SIRST}
\end{figure*}

\begin{figure*}[t]
    \centering
    \includegraphics[width=\textwidth]{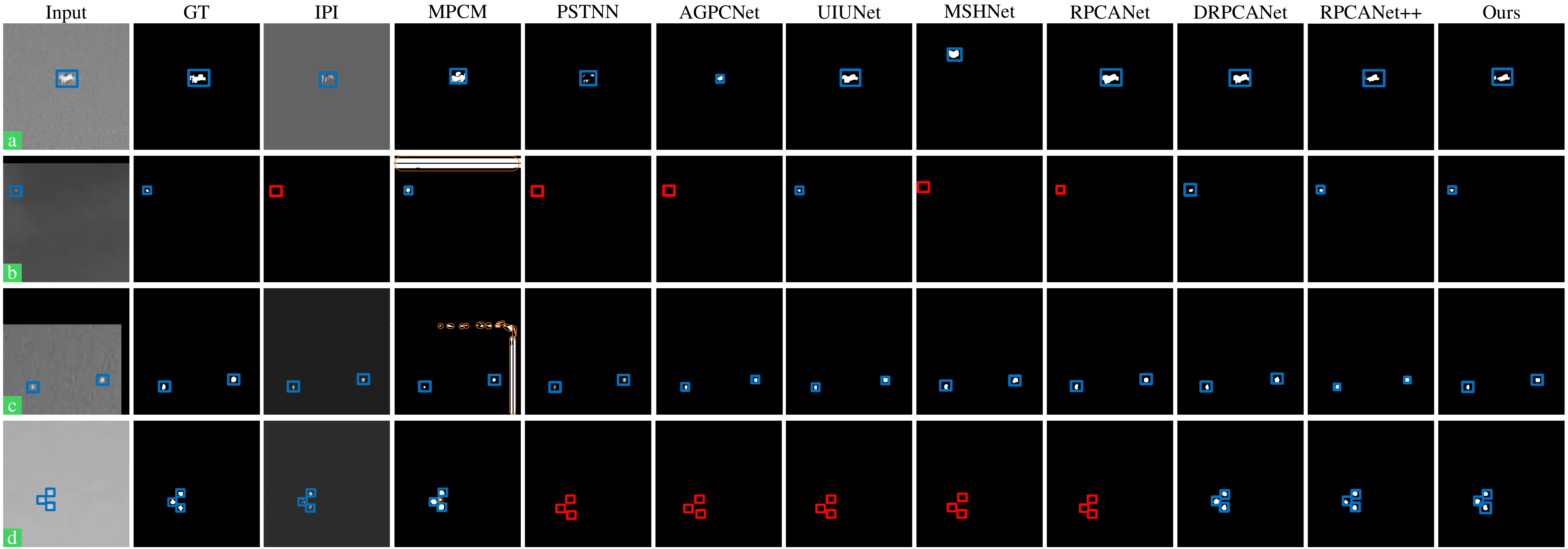}
    \caption{Representative visual results on SIRST-Aug. There are only one target in (a)-(b), two targets in (c), and three targets in (d). The targets are represented by blue, yellow, and red, indicating true positive targets, false positive targets, and false negative targets, respectively.}
\label{Representative_visual_results_of_synthetic_dataset_SIRST-Aug}
\end{figure*}

\begin{figure*}[!h]
    \centering
    \includegraphics[width=\textwidth]{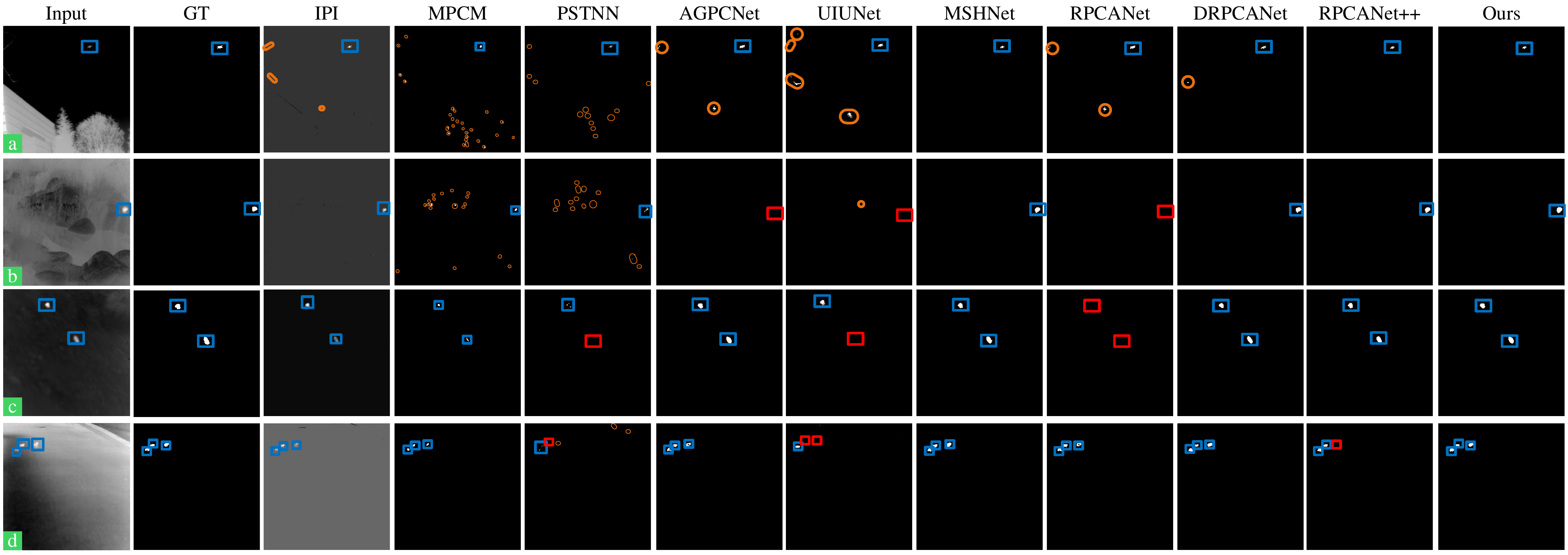}   
    \caption{Representative visual results on IRSTD-1k. There are only one target in (a)-(b), two targets in (c), and three targets in (d). The targets are represented by blue, yellow, and red, indicating true positive targets, false positive targets, and false negative targets, respectively.}
\label{Representative_visual_results_of_real_dataset_IRSTD-1k}
\end{figure*}

\subsubsection{Evaluation Metrics}
As suggested in \cite{wu2024rpcanet}, four metrics are considered to evaluate the performance of target segmentation and infrared image reconstruction.

\begin{itemize}
    \item Mean intersection over union (mIoU) is a pixel-level evaluation metric in semantic segmentation. Let $M$ and $\textrm{IoU}_c$ be the total number of categories and the intersection over union on category $c$. Then mIoU is defined as
\begin{equation}\label{Eq.(18)}
\textrm{mIoU} = \frac{1}{M} \sum_{c=1}^{M} \textrm{IoU}_c.
\end{equation}
    \item $\mathrm{F_1}$-score is the harmonic mean of precision and recall. Precision is the proportion of true positive pixels among the predicted positive pixels, and recall is the proportion of all true positive pixels that have been successfully identified.  $\mathrm{F_1}$ is defined as
\begin{equation}    \label{Eq.(19)}
    \mathrm{F_1} = 2 \cdot \frac{\textrm{precision} \cdot \textrm{recall}}{\textrm{precision} + \textrm{recall}}.
\end{equation}
    \item Probability of detection ($\mathrm{P_d}$) evaluates the proportion of targets correctly detected by the target detection model among all real targets. TP, FP and FN are the same as in (\ref{Eq.(17)}). $\mathrm{P_d}$ is defined as
\begin{equation}
    \mathrm{P_d} = \frac{\textrm{TP}}{\textrm{TP} + \textrm{FN}}.
    \label{Eq.(20)}
\end{equation}

    \item False alarm rate ($\mathrm{F_a}$)  measures the ratio of falsely predicted pixels ($\textrm{P}_{\textrm{false}}$) also in all image pixels ($\textrm{P}_{\textrm{all}}$), which is defined as
\begin{equation}
    \mathrm{F_a} = \frac {\textrm{P}_{\textrm{false}}}{\textrm{P}_{\textrm{all}}}.
    \label{Eq.(21)}
\end{equation}
\end{itemize}

Therefore, mIoU is adopted to evaluate target segmentation and others to evaluate image reconstruction.

\begin{table*}[t]
    \centering
    \renewcommand{\arraystretch}{1.2} 
    \caption{Ablation studies on the SENets. The best results are marked in \textbf{\textcolor{red}{red}}.}
    \label{Ablation_studies_on_the_SENets}
    \resizebox{\textwidth}{!}{
    \begin{tabular}{|c|c|c|c|c|c|c|c|c|c|c|c|c|c|c|c|c|}
    \hline
    \multicolumn{4}{|c|}{SENets} & \multicolumn{4}{c|}{NUDT-SIRST} & \multicolumn{4}{c|}{SIRST-Aug} & \multicolumn{4}{c|}{IRSTD-1k} \\ \cline{1-16}
    SEBEM & SETEM & SENRM & SEIRM & mIoU $\uparrow$ & $\mathrm{F_1}$ $\uparrow$ & $\mathrm{P_d}$ $\uparrow$ & $\mathrm{F_a}$ $\downarrow$ & mIoU $\uparrow$ & $\mathrm{F_1}$ $\uparrow$ & $\mathrm{P_d}$ $\uparrow$ & $\mathrm{F_a}$ $\downarrow$ & mIoU $\uparrow$ & $\mathrm{F_1}$ $\uparrow$ & $\mathrm{P_d}$ $\uparrow$ & $\mathrm{F_a}$ $\downarrow$ \\
    \hline   \hline
    \rowcolor{gray!10} \ding{55} & \ding{55} & \ding{55} & \ding{55} & 73.56 & 78.45 & 79.36 & 8.56  & 60.75 & 70.28 & 81.24 & 43.67 & 50.26 & 61.34 & 70.57 & 15.95 \\ 
     \ding{51} & \ding{55} & \ding{55} & \ding{55} & 80.57 & 81.39 & 86.35 & 5.68  & 65.96 & 75.37 & 86.99 & 39.82 & 55.84 & 65.26 & 76.36 & 11.12 \\ 
    \rowcolor{gray!10} \ding{51} & \ding{51} & \ding{55} & \ding{55} & 88.36 & 89.45 & 90.18 & 4.00  & 70.17 & 80.78 & 93.17 & 34.78 & 60.56 & 72.58 & 83.70 & 8.10 \\ 
    \ding{51} & \ding{51} & \ding{51} & \ding{55} & 91.14 & 96.06 & 97.18 & 2.05 & 73.27 & 84.45 & 98.07 & 27.73 & 63.58 & 77.53 & 88.71 & 3.95 \\ 
    \rowcolor{blue!10} \ding{51} & \ding{51} & \ding{51} & \ding{51} & \textbf{\textcolor{red}{92.37}} & \textbf{\textcolor{red}{96.54}} & \textbf{\textcolor{red}{98.41}} & \textbf{\textcolor{red}{1.79}}  & \textbf{\textcolor{red}{74.56}} & \textbf{\textcolor{red}{85.43}} & \textbf{\textcolor{red}{99.17}} & \textbf{\textcolor{red}{29.78}} & \textbf{\textcolor{red}{64.68}} & \textbf{\textcolor{red}{78.55}} & \textbf{\textcolor{red}{89.39}} & \textbf{\textcolor{red}{4.66}} \\ 
    \hline
    \end{tabular}
    }
\end{table*}

\begin{figure*}[!h]
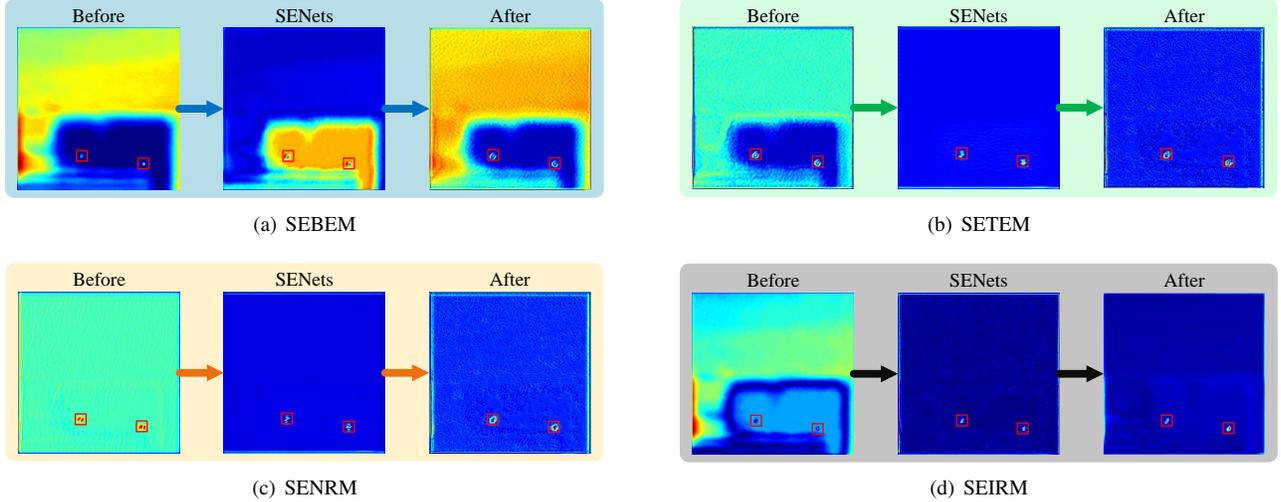

    \centering
    \subfigure[SEBEM]{
    \centering
    \includegraphics[width=0.9\columnwidth]{Figures/Feature_maps_of_SENets_SEBEM.pdf}
    }
    \hspace{0.5cm}
    \subfigure[SETEM]{
    \centering
    \includegraphics[width=0.9\columnwidth]{Figures/Feature_maps_of_SENets_SETEM.pdf}
    }
    \subfigure[SENRM]{
    \centering
    \includegraphics[width=0.9\columnwidth]{Figures/Feature_maps_of_SENets_SENRM.pdf}
    }
    \hspace{0.5cm}
    \subfigure[SEIRM]{
    \centering
    \includegraphics[width=0.9\columnwidth]{Figures/Feature_maps_of_SENets_SEIRM.pdf}
    }
    \caption{Visual feature maps of the SENets operations of each module. ``Before" represents the input feature maps. ``After" denotes the output feature maps.}
    \label{The_feature_maps_of_SENets}
\end{figure*}

\subsection{Experimental Comparisons} \label{sec-b}

\subsubsection{Quantitative Results}

Table \ref{performance_comparison} lists the experimental results of all compared methods, with the best method highlighted in \textbf{\textcolor{red}{red}}. Clearly, compared to data-driven and model-data-driven methods, these model-based methods, \textit{i.e.}, IPI, MPCM, and PSTNN, perform poorly in termd of mIoU and $\mathrm{F_1}$, with a significant gap. This demonstrates that the introduction of neural networks improves semantic segmentation capabilities. Furthermore, data-driven methods (such as AGPCNet) suffer from overfitting and parameter issues, leading to unstable performance across different datasets. In contrast, model-data-driven methods successfully complete the ISTD task by combining physics-based prior knowledge with data-guided accurate targets extraction and segmentation. Note that RPCANet lacks robustness due to its lack of exploration of channel mapping relationships, while RPCANet++ requires more network parameters due to its complex network structure design. In general, our proposed  L-RPCANet achieves satisfactory performance by constructing a hierarchical channel mapping structure with attention mechanisms and introducing a noise reduction module. It also has the advantages of fewer parameters and faster GPU inference time, see Fig. \ref{fig:Time-mIoU}.

Furthermore, Table \ref{AUC_comparison_of_different_methods} provides the related area under the receiver operating characteristic (ROC) curve results. It is seen that our proposed  L-RPCANet maintains a high AUC on almost all datasets, demonstrating its outstanding cross-domain robustness.

\begin{table*}[t]
    \centering
    \setlength{\tabcolsep}{6pt} 
    \renewcommand{\arraystretch}{1.2} 
    \caption{Effects of stage number K on the detection performance. The best results are marked in \textbf{\textcolor{red}{red}}.}
    \label{Stage_K_effect}
    \begin{tabular}{|c|c|c|c|c|c|c|c|c|c|c|c|c|c|}
    \hline
    \multirow{2}{*}{$K$} & \multirow{2}{*}{Params} & \multicolumn{4}{c|}{NUDT-SIRST} & \multicolumn{4}{c|}{SIRST-Aug} & \multicolumn{4}{c|}{IRSTD-1k}  \\ \cline{3-14}
 & & mIoU $\uparrow$ & $\mathrm{F_1}$ $\uparrow$ & $\mathrm{P_d}$ $\uparrow$ & $\mathrm{F_a}$ $\downarrow$ & mIoU $\uparrow$ & $\mathrm{F_1}$ $\uparrow$ & $\mathrm{P_d}$ $\uparrow$ & $\mathrm{F_a}$ $\downarrow$ & mIoU $\uparrow$ & $\mathrm{F_1}$ $\uparrow$ & $\mathrm{P_d}$ $\uparrow$ & $\mathrm{F_a}$ $\downarrow$ \\
    \hline   \hline
        \rowcolor{gray!10} 1 & 0.0360M  & 91.39 & 95.51 & 98.52 & 2.16 & 72.83 & 84.28 & 98.07 & 29.14 & 61.26 & 75.98 & 92.12 & 6.29 \\ 
        2 & 0.0720M  & 91.61 & 95.62 & 97.88 & 2.08  & 72.96 & 84.37 & 98.9 & 34.82 & 61.59 & 76.24 & 86.99 & 5.12 \\ 
        \rowcolor{gray!10} 3 & 0.1080M  & 91.53 & 95.58 & 97.98 & 2.00  & 73.58 & 84.78 & 99.17 & 32.83 & 62.60 & 77.00 & 88.7 & 5.21 \\ 
        4 & 0.1439M  & 90.06 & 95.06 & 97.88 & 1.95 & 73.81 & 84.93 & 98.21 & 28.51 &  61.98 & 76.53 & 87.33 & \textbf{\textcolor{red}{4.35}} \\ 
        \rowcolor{gray!10} 5 & 0.1799M  & 91.64 & 95.64 & 98.31 & 2.01  & 74.28 & 85.24 & 98.76 & 28.06 & 63.45 & 77.63 & 88.36 & 5.45 \\ 
        \rowcolor{blue!10} 6 & 0.216M & \textbf{\textcolor{red}{92.37}} & \textbf{\textcolor{red}{96.04}} & \textbf{\textcolor{red}{98.41}} & \textbf{\textcolor{red}{1.79}} & \textbf{\textcolor{red}{74.56}} & \textbf{\textcolor{red}{85.43}}  & \textbf{\textcolor{red}{99.17}} & 29.78 &  \textbf{\textcolor{red}{64.68}} & \textbf{\textcolor{red}{78.55}} & \textbf{\textcolor{red}{89.39}} & 4.66 \\         
        7 & 0.2519M  & 88.53 & 93.58 & 96.77 &3.74 &72.29 & 83.92 & 98.9 & \textbf{\textcolor{red}{27.38}} & 62.29 & 76.76 & 87.33 & 4.82 \\ 
    \hline
    \end{tabular}
\end{table*}

\subsubsection{Qualitative Results}

To visualize the detection performance, Fig. \ref{Representative_visual_results_of_synthetic_dataset_NUDT-SIRST} plots the detection results on NUDT-SIRST, where ``Input" represents the original image and ``GT" represents the reference ground truth image. It is found that model-based methods suffer extremely high false positives and false negatives due to limited feature extraction from the image (a). This indicates that the ISTD task in real-world scenarios is difficult to be handled by simple low-rank and sparse models. Data-driven methods with automatic feature learning reduce false positive targets, consistent with the quantitative results in Table \ref{performance_comparison}. 
However, the double-layer nested structure adopted by UIUNet retains skip connections, resulting in low features extracted by the convolution kernels in the continuous smooth areas. This causes the network to overwhelm weak targets gradients during backpropagation and thus missed detected targets. Although MSHNet can detect the targets in the cloud background, there is no noise prior introduced in the multiscale feature concatenation stage and no mechanism to suppress high-frequency isolated spikes, leading to the high-contrast noise remaining prominent after fusion. For all testing images in Fig. \ref{Representative_visual_results_of_synthetic_dataset_NUDT-SIRST}, model-data-driven methods can successfully detect small targets, which verifies the effectiveness of the combination of model-based and data-driven methods.

Fig. \ref{Representative_visual_results_of_synthetic_dataset_SIRST-Aug} and Fig. \ref{Representative_visual_results_of_real_dataset_IRSTD-1k} present the detection results for SIRST-Aug and IRSTD-1k, respectively. It is observed that RPCANet exhibits some false positives and false negatives due to the excessive shrinkage of sparse weights in high-variance environments from Fig. \ref{Representative_visual_results_of_real_dataset_IRSTD-1k}. DRPCANet also suffers from false positives in some scenarios due to its over-reliance on input features when generating dynamic parameters. At the same time, the excessive requirements for model self-training in RPACNet++ have led to the occurrence of false negatives. Our proposed method  simultaneously projects the input image into three learnable subspaces and dynamically modifies the weights of each channel through the attention mechanism. This method can adaptively tighten the background distribution and expand the target distribution in different datasets, ultimately suppressing false alarms and missed detections.

\begin{figure}[t]
    \centering
    \subfigure[mIoU]{
        \includegraphics[width=3.5cm]{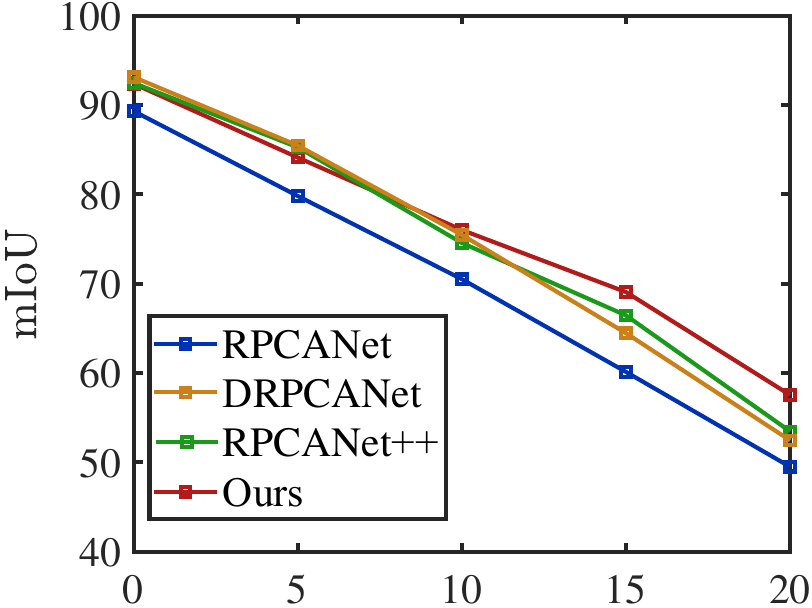}
    }
    \hspace{-0.2cm}
    \subfigure[$\mathrm{F_1}$]{
        \includegraphics[width=3.5cm]{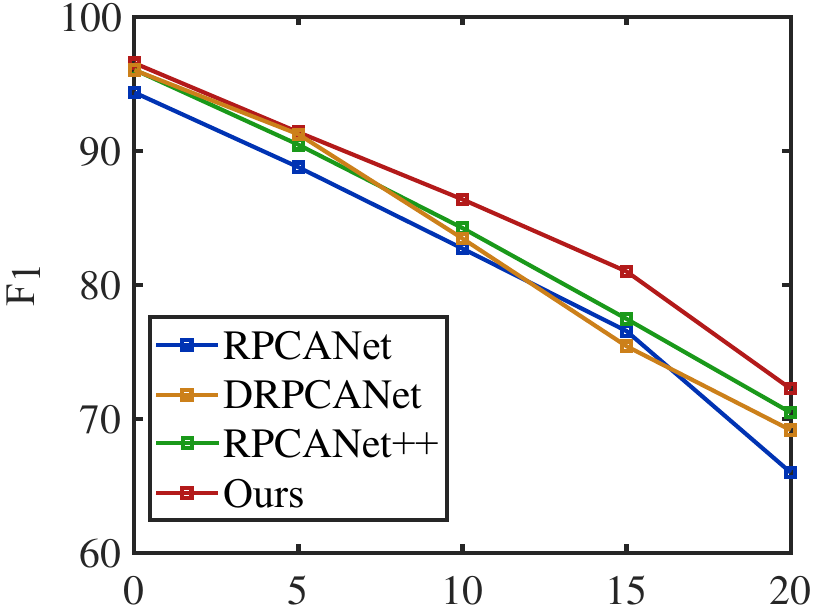}
    }

    \subfigure[$\mathrm{P_d}$]{
        \includegraphics[width=3.5cm]{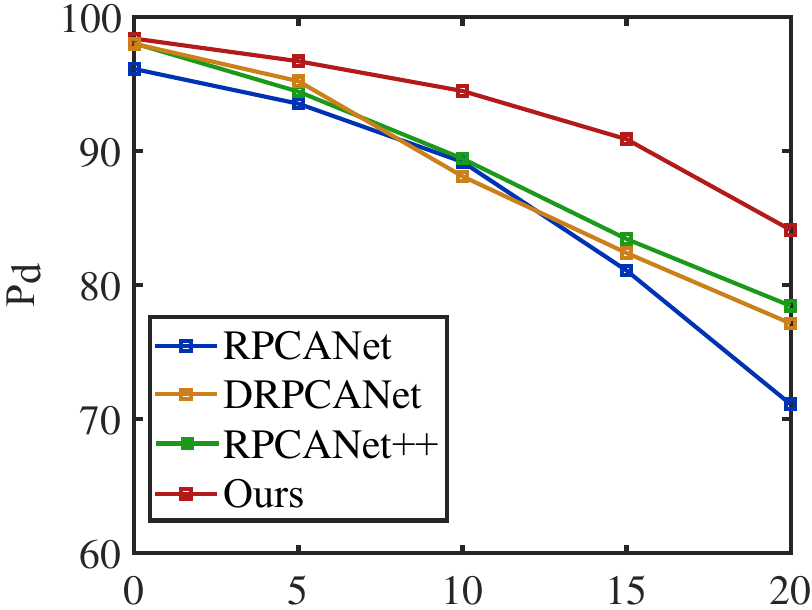}
    }
    \hspace{-0.2cm}
    \subfigure[$\mathrm{F_a}$]{
        \includegraphics[width=3.5cm]{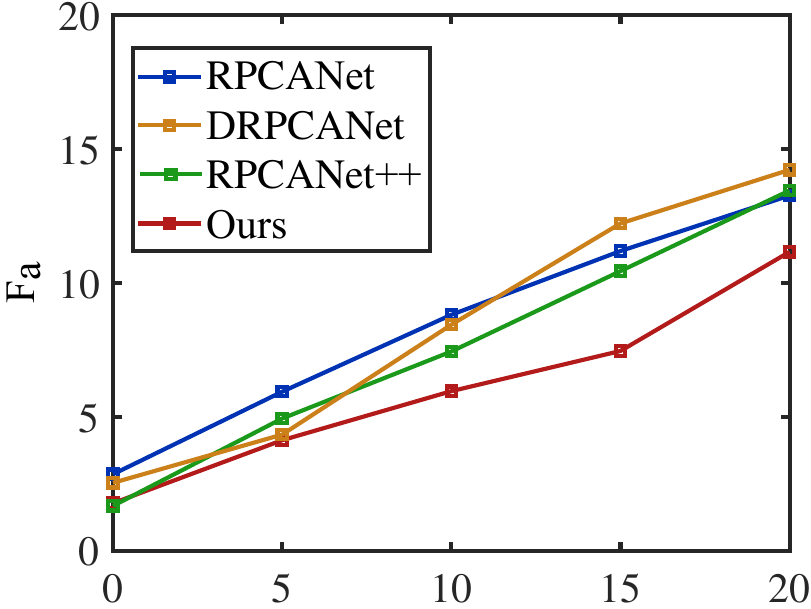}
    }
        \caption{Results of background clutter under Gaussian noise with mean 0 and variance $[0, 5, 10, 15, 20]$ on NUDT-SIRST.}
    \label{Robustness_analysis_of_gaussian_noise}

\end{figure}

\subsection{Ablation Studies} \label{sec-c}

As shown in Table \ref{Ablation_studies_on_the_SENets}, SENets contribute to enhancing the detection performance in each module. The extraction of the targets in \texttt{SETEM} and the reduction of noise in \texttt{SENRM} depend on the optimization of the channel weight for the targets and the noise features, which is the operation of SENets correspondingly. For the complex background estimation task handled by \texttt{SEBEM}, it is based on both the adjustment of channel weights and spatial features. Therefore, after incorporating SENets into \texttt{SEBEM}, the detection performance has achieved a relatively better improvement, but it is not as significant as that of \texttt{SETEM} and \texttt{SENRM}. In contrast, the image reconstruction of \texttt{SEIRM} is based on the precise restoration of spatial information, and the performance improvement of SENets in the image reconstruction module is relatively less.

Furthermore,  Fig. \ref{The_feature_maps_of_SENets} visualizes the features extracted by each module in the first stage to verify the contributions. Here, \texttt{SEBEM} first estimates the shape and structure of the targets in the image background, \texttt{SETEM} extracts possible targets, \texttt{SENRM} obtains combined noise information that belongs neither to the background nor to the targets, and \texttt{SEIRM} reconstructs an image that contains only the target information based on the information obtained from the previous three modules. Due to the application of double-layer channel feature connections for each module and the intervention of the noise reduction module, our proposed method  accurately identifies background structures and obtains the targets with fewer iterations.

\begin{figure}[t]
    \centering
    \subfigure[mIoU]{
        \includegraphics[width=3.5cm]{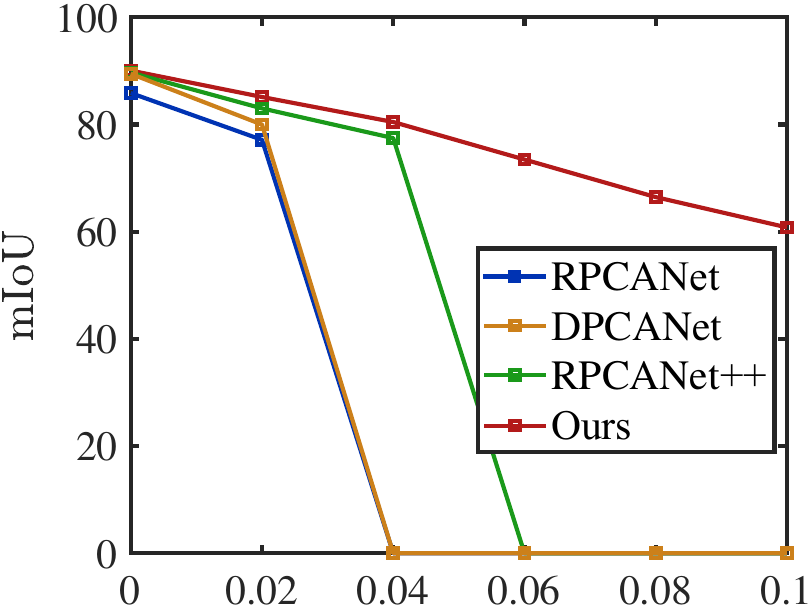}
    }
    \hspace{-0.2cm}
    \subfigure[$\mathrm{F_1}$]{
        \includegraphics[width=3.5cm]{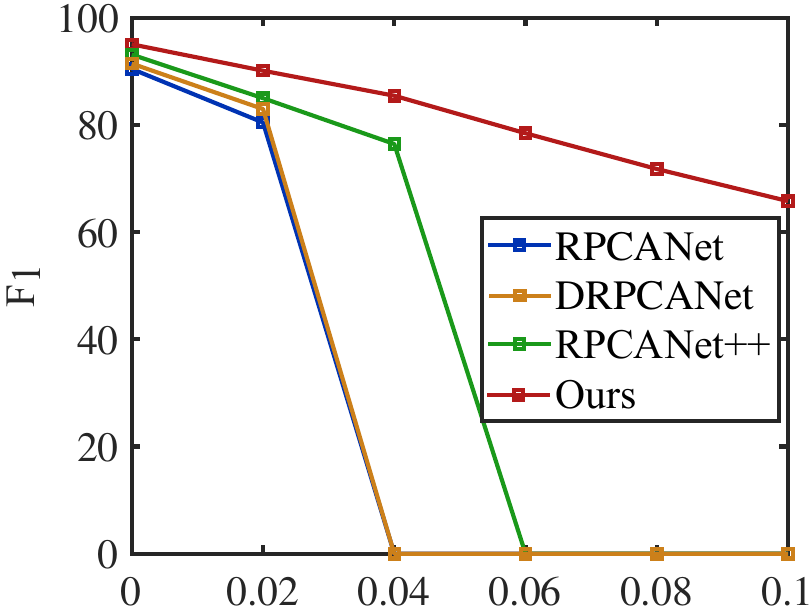}
    }

    \subfigure[$\mathrm{P_d}$]{
        \includegraphics[width=3.5cm]{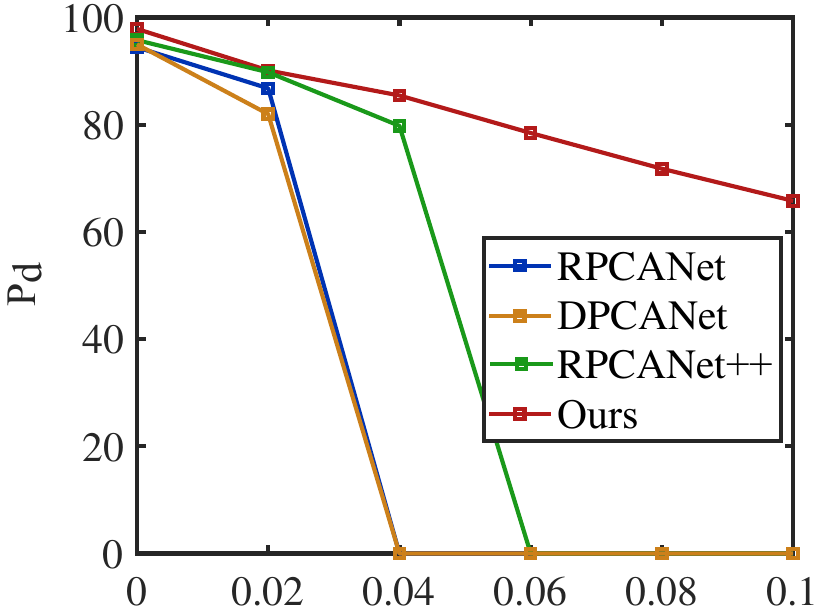}
    }
    \hspace{-0.2cm}
    \subfigure[$\mathrm{F_a}$]{
        \includegraphics[width=3.5cm]{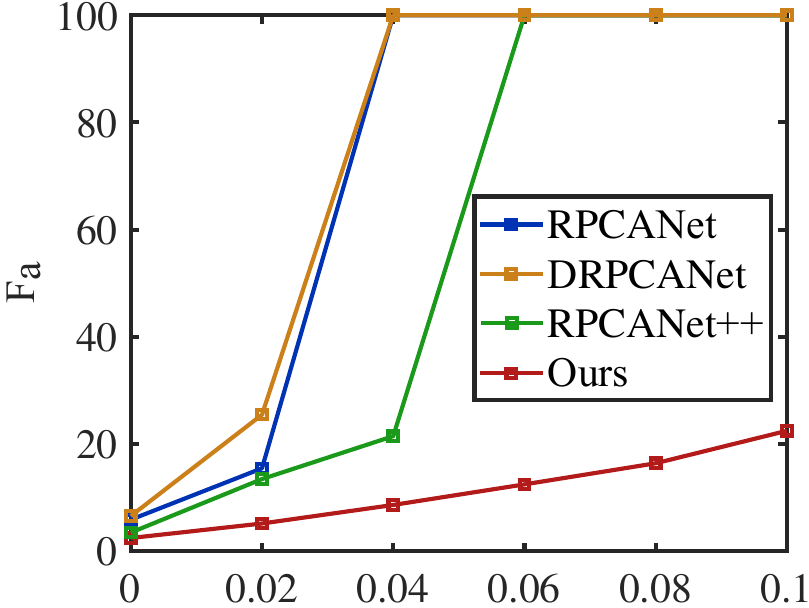}
    }
            \caption{Results of foreground noise under salt-and-pepper noise with salt $[0, 0.02, 0.04, 0.06, 0.08, 0.10]$ and pepper $0.04$ on NUDT-SIRST.} 
    \label{Robustness_analysis_of_foreground_noise}
\end{figure}

\begin{table*}[t]
    \centering
    \setlength{\tabcolsep}{6pt} 
    \renewcommand{\arraystretch}{1.2} 
    \caption{Effects of channel numbers of intermediate bottleneck layers ($BC$) and and total channel number ($C$) on the detection performance. The best results are marked in \textbf{\textcolor{red}{red}}.}
    \label{BC_effect_and_C_effect}
    \begin{tabular}{|c|c|c|c|c|c|c|c|c|c|c|c|c|c|c|}
    \hline
    \multirow{2}{*}{Layers} & \multirow{2}{*}{Channels} & \multicolumn{4}{c|}{NUDT-SIRST} & \multicolumn{4}{c|}{SIRST-Aug} & \multicolumn{4}{c|}{IRSTD-1k} \\ \cline{3-14}
 & & mIoU $\uparrow$ & $\mathrm{F_1}$ $\uparrow$ & $\mathrm{P_d}$ $\uparrow$ & $\mathrm{F_a}$ $\downarrow$ & mIoU $\uparrow$ & $\mathrm{F_1}$ $\uparrow$ & $\mathrm{P_d}$ $\uparrow$ & $\mathrm{F_a}$ $\downarrow$ & mIoU $\uparrow$ & $\mathrm{F_1}$ $\uparrow$ & $\mathrm{P_d}$ $\uparrow$ & $\mathrm{F_a}$ $\downarrow$ \\
    \hline   \hline
        \rowcolor{blue!10} \textit{BC} = {4} & \textit{C} = {32}  & \textbf{\textcolor{red}{92.37}} & \textbf{\textcolor{red}{96.04}} & \textbf{\textcolor{red}{98.41}} & \textbf{\textcolor{red}{1.79}}  & \textbf{\textcolor{red}{74.56}} & \textbf{\textcolor{red}{85.43}} & \textbf{\textcolor{red}{99.17}} & \textbf{\textcolor{red}{29.78}} & \textbf{\textcolor{red}{64.68}} & \textbf{\textcolor{red}{78.55}} & 89.39 & \textbf{\textcolor{red}{4.66}}  \\ 
        $BC = 8$ & $C = 32$  & 91.03 & 95.58 & 97.78 & 2.00 & 72.71 & 84.21 & 98.35 & 29.45 & 62.63 & 77.02 & 90.75 & 5.59  \\ 
        \rowcolor{gray!10} $BC = 16$ & $C = 32$ & 90.52 & 94.17 & 96.54 & 2.99 & 72.31 & 83.93 & 97.66 & 27.7 & 62.58 & 76.98 & 88.01 & 4.76  \\ 
        $BC = 4 $ & $C = 40$ & 90.06 & 94.06 & 97.38 & 2.34  & 72.54 & 83.08 & 97.21 & 30.23 & 62.60 & 77.00 & 88.7 & 5.21 \\ 
        \rowcolor{gray!10} $BC = 4 $ & $C = 48$ & 89.34 & 93.34 & 97.00 & 2.54  & 72.1	& 84.03 & 97.32 & 29.33  & 61.16 & 75.90 & 88.7 & 5.13 \\ 
        $BC = 4 $ & $C = 56$ & 89.09 & 92.85 & 97.31 & 3.01  & 70.34 & 83.78 & 96.39 & 31.14 & 60.98 & 75.76 & 91.44 & 5.99 \\ 
        \rowcolor{gray!10} $BC = 4 $ & $C = 64$ & 87.93 & 91.53 & 95.34 & 4.45  & 68.99 & 81.45 & 94.55 & 33.74 & 58.25 & 73.61 & \textbf{\textcolor{red}{92.81}} & 5.84 \\         
    \hline
    \end{tabular}
\end{table*}

\begin{table*}[t]
    \centering
    \setlength{\tabcolsep}{6pt} 
    \renewcommand{\arraystretch}{1.2} 
    \caption{Effects of loss weight $\eta$. The best results are marked in \textbf{\textcolor{red}{red}}.}
    \label{Ablation_studies_on_the_eta}
    \begin{tabular}{|c|c|c|c|c|c|c|c|c|c|c|c|c|}
    \hline
    \multirow{2}{*}{$\eta$} & \multicolumn{4}{c|}{NUDT-SIRST} & \multicolumn{4}{c|}{SIRST-Aug} & \multicolumn{4}{c|}{IRSTD-1k} \\ \cline{2-13}
    & mIoU $\uparrow$ & $\mathrm{F_1}$ $\uparrow$ & $\mathrm{P_d}$ $\uparrow$ & $\mathrm{F_a}$ $\downarrow$ & mIoU $\uparrow$ & $\mathrm{F_1}$ $\uparrow$ & $\mathrm{P_d}$ $\uparrow$ & $\mathrm{F_a}$ $\downarrow$ & mIoU $\uparrow$ & $\mathrm{F_1}$ $\uparrow$ & $\mathrm{P_d}$ $\uparrow$ & $\mathrm{F_a}$ $\downarrow$ \\
    \hline   \hline
    \rowcolor{gray!10} 0.005 & 77.56 & 82.19 & 84.25 & 8.78  & 65.47 & 75.58 & 87.39 & 35.73 & 57.45 & 68.19 & 78.43 & 20.54 \\ 
    \rowcolor{blue!10} 0.01 & \textbf{\textcolor{red}{92.37}} & \textbf{\textcolor{red}{96.54}} & \textbf{\textcolor{red}{98.41}} & \textbf{\textcolor{red}{1.79}}  & \textbf{\textcolor{red}{74.56}} & \textbf{\textcolor{red}{85.43}} & \textbf{\textcolor{red}{99.17}} & \textbf{\textcolor{red}{29.78}} & \textbf{\textcolor{red}{64.68}} & \textbf{\textcolor{red}{78.55}} & \textbf{\textcolor{red}{89.39}} & \textbf{\textcolor{red}{4.66}} \\ 
    \rowcolor{gray!10} 0.015 & 90.36 & 93.45 & 94.18 & 2.90  & 71.17 & 82.78 & 95.17 & 30.78 & 61.56 & 75.58 & 86.70 & 6.10 \\ 
    0.2 & 73.27 & 78.15 & 70.35 & 18.05 & 60.28 & 74.45 & 80.07 & 40.73 & 50.36 & 70.37 & 80.37 & 16.78 \\ 
    \hline
    \end{tabular}
\end{table*}

\begin{figure*}[t]
    \centering
    \subfigure[ NUDT-SIRST ]{
        \includegraphics[width=4.5cm]{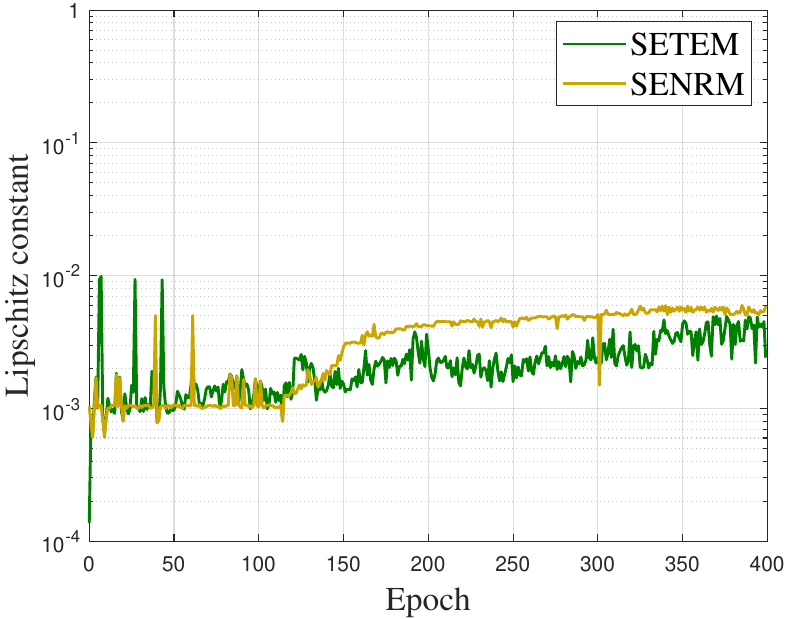}
    }
    \hspace{0.2cm}
    \subfigure[ SIRST-Aug ]{
        \includegraphics[width=4.5cm]{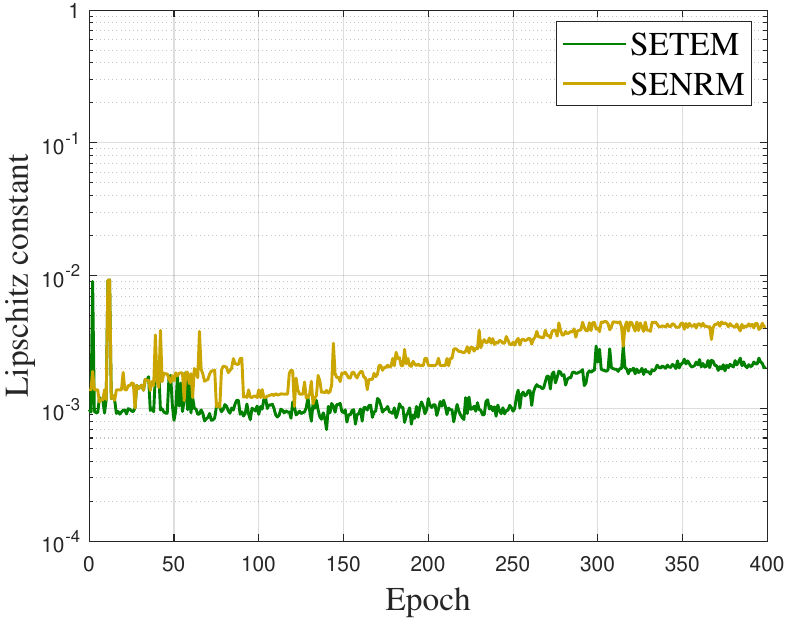}
    }
    \hspace{0.2cm}
    \subfigure[ IRSTD-1k ]{
        \includegraphics[width=4.5cm]{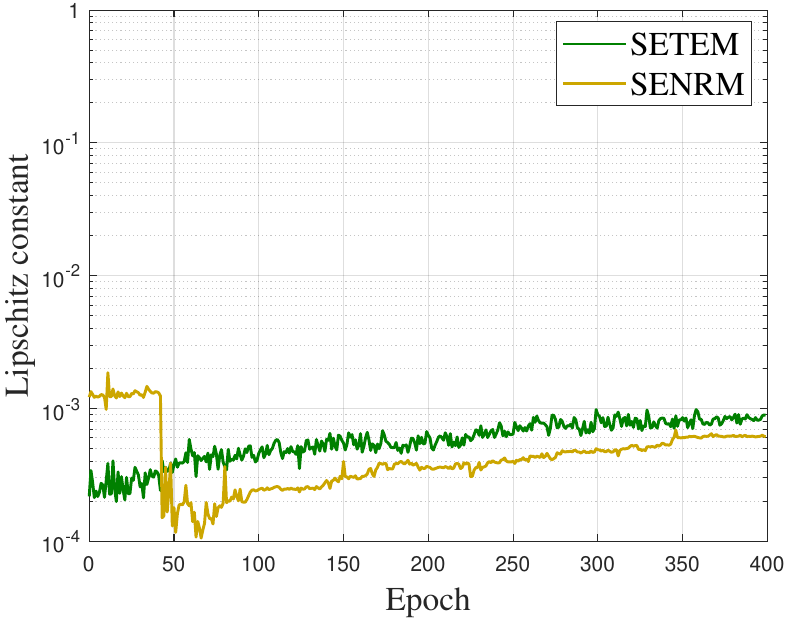}
    }
    \caption{Lipschitz conditions in the \texttt{SETEM} and \texttt{SENRM} that increases with the training epoch.}
    \label{The_Lipschitz_constant_in_the_SETEM_and_SENRM_module}
\end{figure*}


\subsection{Discussions} \label{sec-d}

\subsubsection{Robustness Verification} 

To investigate the robustness, Gaussian noise of different intensities is applied to NUDT-SIRST, which commonly disturbs the background textures of infrared images. Compared to current model-data-driven methods including RPCANet, DRPCANet, and RPCANet++, it is concluded from Fig. \ref{Robustness_analysis_of_gaussian_noise} that the detection performance of all methods decreases with increasing noise intensity, but our proposed method has a smaller decrease rate. When noise is applied that affects the intensity of infrared small targets, such as salt-and-pepper noise, RPCANet, DRPCANet, and RPCANet++ lose their detection ability at a certain noise intensity, while our proposed method  still has good detection performance, as seen in Fig. \ref{Robustness_analysis_of_foreground_noise}. In conclusion, quantitative results show that our proposed method is robust to both background clutter and foreground noise.

\subsubsection{Effects of Parameter $K$}
Table \ref{Stage_K_effect} illustrates the effects of different numbers of image decomposition stages. As the number of $K$ increases, the detection performance improves, but the size of the parameters also increases. In addition, when $K$ is 7, the performance is significantly worse than that of $K = 6$, which makes sense that a higher number of stages $K$ will cause excessive growth of the gradient and thus hinder its propagation.
Therefore, $K$ is set to 6 in the experiments.

\subsubsection{Effects of Channel Numbers} 

Table \ref{BC_effect_and_C_effect} discusses the difference between different channels in layers ($BC$) and the total channels ($C$), which denote the channel number of the bottleneck layers and the total channel number in the second channel mapping of the networks. Obviously, the detection performance is the best when $BC = 4$ and $C = 32$. This is because the number of channels in the intermediate bottleneck layers $BC$ is too large, and assigning the extracted features to the convolution layer with a channel number of $C$ results in a reduced mapping relationship. When $C$ is too large, the features of small targets can be divided too dispersively, leading to a sharp performance decline in the image reconstruction stage. Therefore, $BC$ and $C$ are respectively set to 4 and 32.

\subsubsection{Effects of Loss Weight $\eta$}
Table \ref{Ablation_studies_on_the_eta} illustrates that when $\eta = 0.01$, it achieves the best detection performance in the NUDT-SIRST, SIRST-Aug, and IRSTD-1k datasets. A larger value (say 0.2) may overemphasize the reconstruction, while a smaller value (say 0.005) may reduce the regularization. Based on this observation, $\eta = 0.01$ is set as the default setting.

\subsubsection{Lipschitz Conditions} 
As shown in Fig. \ref{The_Lipschitz_constant_in_the_SETEM_and_SENRM_module}, the Lipschitz constant gradually tends towards a stable value as training progresses in three datasets. This not only validates that $\mathcal{S}(\mathbf{T})$ and $\mathcal{G}(\mathbf{N})$ are convergent, but also satisfies the assumption of convergence in these modules.

\section{Conclusion}\label{Conclusion}

In this paper, we have proposed a lightweight, robust, and interpretable ISTD framework with channel attention mechanisms. The architecture has four network modules, including proximal networks to estimate the background, sparse constrained networks to extract the targets, noise reduction neural layers to reduce noise, and simple CNNs to reconstruct the image. 
Our scheme has achieved reliable visualizations and excellent detection results in numerous experiments on various datasets, in which the mIoU has increased at least by 3\% and $\mathrm{F_1}$ has increased by 2\% compared to existing methods. Our architecture effectively drives the combination of neural layers and channel attention mechanism to learn low-rank backgrounds, sparse targets, noise reduction, and image reconstruction, almost removing the ``black-box" nature of the neural network in detection tasks.

In the future, we are interested in developing a framework that can adaptively select stages and channel numbers. Besides, integrating the proposed method with state-space models is also a promising research direction.

\bibliographystyle{IEEEtran}
\bibliography{IEEEabrv,mybibfile}

\begin{thebibliography}{10}
\providecommand{\url}[1]{#1}
\csname url@samestyle\endcsname
\providecommand{\newblock}{\relax}
\providecommand{\bibinfo}[2]{#2}
\providecommand{\BIBentrySTDinterwordspacing}{\spaceskip=0pt\relax}
\providecommand{\BIBentryALTinterwordstretchfactor}{4}
\providecommand{\BIBentryALTinterwordspacing}{\spaceskip=\fontdimen2\font plus
\BIBentryALTinterwordstretchfactor\fontdimen3\font minus \fontdimen4\font\relax}
\providecommand{\BIBforeignlanguage}[2]{{%
\expandafter\ifx\csname l@#1\endcsname\relax
\typeout{** WARNING: IEEEtran.bst: No hyphenation pattern has been}%
\typeout{** loaded for the language `#1'. Using the pattern for}%
\typeout{** the default language instead.}%
\else
\language=\csname l@#1\endcsname
\fi
#2}}
\providecommand{\BIBdecl}{\relax}
\BIBdecl

\bibitem{yue2023dif}
J.~Yue, L.~Fang, S.~Xia, Y.~Deng, and J.~Ma, ``Dif-fusion: Toward high color fidelity in infrared and visible image fusion with diffusion models,'' \emph{IEEE Transactions on Image Processing}, vol.~32, pp. 5705--5720, 2023.

\bibitem{li2023lrrnet}
H.~Li, T.~Xu, X.-J. Wu, J.~Lu, and J.~Kittler, ``{LRRNet}: A novel representation learning guided fusion network for infrared and visible images,'' \emph{IEEE Transactions on Pattern Analysis and Machine Intelligence}, vol.~45, no.~9, pp. 11\,040--11\,052, 2023.

\bibitem{lin2024learning}
F.~Lin, S.~Ge, K.~Bao, C.~Yan, and D.~Zeng, ``Learning shape-biased representations for infrared small target detection,'' \emph{IEEE Transactions on Multimedia}, vol.~26, pp. 4681--4692, 2024.

\bibitem{yuan2024sctransnet}
S.~Yuan, H.~Qin, X.~Yan, N.~Akhtar, and A.~Mian, ``{SCT}rans{N}et: Spatial-channel cross transformer network for infrared small target detection,'' \emph{IEEE Transactions on Geoscience and Remote Sensing}, vol.~62, pp. 1--15, 2024.

\bibitem{wang2024vivo}
F.~Wang, Y.~Zhong, O.~Bruns, Y.~Liang, and H.~Dai, ``In vivo {NIR-II} fluorescence imaging for biology and medicine,'' \emph{Nature Photonics}, vol.~18, no.~6, pp. 535--547, 2024.

\bibitem{liu2025enhancing}
Z.~Liu, Y.~Zhang, J.~He, T.~Zhang, S.~ur~Rehman, M.~Saraee, and C.~Sun, ``Enhancing infrared small target detection: A saliency-guided multi-task learning approach,'' \emph{IEEE Transactions on Intelligent Transportation Systems}, vol.~26, no.~3, pp. 3603--3618, 2025.

\bibitem{li2025ilnet}
H.~Li, J.~Yang, R.~Wang, and Y.~Xu, ``{ILNet}: Low-level matters for salient infrared small target detection,'' \emph{IEEE Transactions on Aerospace and Electronic Systems}, vol.~61, no.~4, pp. 8306--8318, 2025.

\bibitem{liu2025graph}
T.~Liu, Y.~Liu, J.~Yang, B.~Li, Y.~Wang, and W.~An, ``Graph {L}aplacian regularization for fast infrared small target detection,'' \emph{Pattern Recognition}, vol. 158, p. 111077, 2025.

\bibitem{zhao2022single}
M.~Zhao, W.~Li, L.~Li, J.~Hu, P.~Ma, and R.~Tao, ``Single-frame infrared small-target detection: A survey,'' \emph{IEEE Geoscience and Remote Sensing Magazine}, vol.~10, no.~2, pp. 87--119, 2022.

\bibitem{kumar2025small}
N.~Kumar and P.~Singh, ``Small and dim target detection in infrared imagery: A review, current techniques and future directions,'' \emph{Neurocomputing}, p. 129640, 2025.

\bibitem{gao2018infrared}
C.~Gao, L.~Wang, Y.~Xiao, Q.~Zhao, and D.~Meng, ``Infrared small-dim target detection based on markov random field guided noise modeling,'' \emph{Pattern Recognition}, vol.~76, pp. 463--475, 2018.

\bibitem{zhang2025m4net}
F.~Zhang, H.~Hu, B.~Zou, and M.~Luo, ``{M4Net}: Multi-level multi-patch multi-receptive multi-dimensional attention network for infrared small target detection,'' \emph{Neural Networks}, vol. 183, p. 107026, 2025.

\bibitem{gao2013infrared}
C.~Gao, D.~Meng, Y.~Yang, Y.~Wang, X.~Zhou, and A.~G. Hauptmann, ``Infrared patch-image model for small target detection in a single image,'' \emph{IEEE Transactions on Image Processing}, vol.~22, no.~12, pp. 4996--5009, 2013.

\bibitem{wu2023uiu}
X.~Wu, D.~Hong, and J.~Chanussot, ``{UIU}-{N}et: {U}-{N}et in {U}-{N}et for infrared small object detection,'' \emph{IEEE Transactions on Image Processing}, vol.~32, pp. 364--376, 2023.

\bibitem{wu2024rpcanet}
F.~Wu, T.~Zhang, L.~Li, Y.~Huang, and Z.~Peng, ``{RPCANet}: Deep unfolding {RPCA} based infrared small target detection,'' in \emph{Proceedings of the IEEE/CVF Winter Conference on Applications of Computer Vision}, 2024, pp. 4809--4818.

\bibitem{liu2025tensor}
J.~Liu, M.~Feng, X.~Xiu, X.~Zeng, and J.~Zhang, ``Tensor low-rank approximation via plug-and-play priors for anomaly detection in remote sensing images,'' \emph{IEEE Transactions on Instrumentation and Measurement}, vol.~74, pp. 1--14, 2025.

\bibitem{sun2023learning}
J.~Sun, X.~Xiu, Z.~Luo, and W.~Liu, ``Learning high-order multi-view representation by new tensor canonical correlation analysis,'' \emph{IEEE Transactions on Circuits and Systems for Video Technology}, vol.~33, no.~10, pp. 5645--5654, 2023.

\bibitem{xiu2024efficient}
X.~Xiu, L.~Pan, Y.~Yang, and W.~Liu, ``Efficient and fast joint sparse constrained canonical correlation analysis for fault detection,'' \emph{IEEE Transactions on Neural Networks and Learning Systems}, vol.~35, no.~3, pp. 4153--4163, 2024.

\bibitem{zhu2020tnlrs}
H.~Zhu, H.~Ni, S.~Liu, G.~Xu, and L.~Deng, ``{TNLRS}: Target-aware non-local low-rank modeling with saliency filtering regularization for infrared small target detection,'' \emph{IEEE Transactions on Image Processing}, vol.~29, pp. 9546--9558, 2020.

\bibitem{zhang2019infrared}
L.~Zhang and Z.~Peng, ``Infrared small target detection based on partial sum of the tensor nuclear norm,'' \emph{Remote Sensing}, vol.~11, no.~4, p. 382, 2019.

\bibitem{wei2016multiscale}
Y.~Wei, X.~You, and H.~Li, ``Multiscale patch-based contrast measure for small infrared target detection,'' \emph{Pattern Recognition}, vol.~58, pp. 216--226, 2016.

\bibitem{liu2023combining}
T.~Liu, Q.~Yin, J.~Yang, Y.~Wang, and W.~An, ``Combining deep denoiser and low-rank priors for infrared small target detection,'' \emph{Pattern Recognition}, vol. 135, p. 109184, 2023.

\bibitem{pang2024lrta}
D.~Pang, T.~Shan, Y.~Ma, P.~Ma, T.~Hu, and R.~Tao, ``{LRTA}-{SP}: Low rank tensor approximation with saliency prior for small target detection in infrared videos,'' \emph{IEEE Transactions on Aerospace and Electronic Systems}, vol.~61, no.~2, pp. 2644--2658, 2025.

\bibitem{luo2024revisiting}
Y.~Luo, X.~Zhao, and D.~Meng, ``Revisiting nonlocal self-similarity from continuous representation,'' \emph{IEEE Transactions on Pattern Analysis and Machine Intelligence}, vol.~47, no.~1, pp. 450--468, 2025.

\bibitem{chen2022local}
F.~Chen, C.~Gao, F.~Liu, Y.~Zhao, Y.~Zhou, D.~Meng, and W.~Zuo, ``Local patch network with global attention for infrared small target detection,'' \emph{IEEE Transactions on Aerospace and Electronic Systems}, vol.~58, no.~5, pp. 3979--3991, 2022.

\bibitem{zhang2024irprunedet}
M.~Zhang, H.~Yang, J.~Guo, Y.~Li, X.~Gao, and J.~Zhang, ``{IRP}rune{D}et: efficient infrared small target detection via wavelet structure-regularized soft channel pruning,'' in \emph{Proceedings of the AAAI Conference on Artificial Intelligence}, vol.~38, no.~7, 2024, pp. 7224--7232.

\bibitem{zhang2023attention}
T.~Zhang, L.~Li, S.~Cao, T.~Pu, and Z.~Peng, ``Attention-guided pyramid context networks for detecting infrared small target under complex background,'' \emph{IEEE Transactions on Aerospace and Electronic Systems}, vol.~59, no.~4, pp. 4250--4261, 2023.

\bibitem{he2016deep}
K.~He, X.~Zhang, S.~Ren, and J.~Sun, ``Deep residual learning for image recognition,'' in \emph{Proceedings of the IEEE Conference on Computer Vision and Pattern Recognition}, 2016, pp. 770--778.

\bibitem{liu2024infrared}
Q.~Liu, R.~Liu, B.~Zheng, H.~Wang, and Y.~Fu, ``Infrared small target detection with scale and location sensitivity,'' in \emph{Proceedings of the IEEE/CVF Computer Vision and Pattern Recognition}, 2024.

\bibitem{ronneberger2015u}
O.~Ronneberger, P.~Fischer, and T.~Brox, ``U-net: Convolutional networks for biomedical image segmentation,'' in \emph{International Conference on Medical Image Computing and Computer-Assisted Intervention}.\hskip 1em plus 0.5em minus 0.4em\relax Springer, 2015, pp. 234--241.

\bibitem{yang2025deep}
Z.~Yang, H.~Yu, J.~Zhang, Q.~Tang, and A.~Mian, ``Deep learning based infrared small object segmentation: Challenges and future directions,'' \emph{Information Fusion}, vol. 118, p. 103007, 2025.

\bibitem{wu2024extrapolated}
Z.~Wu, C.~Huang, and T.~Zeng, ``Extrapolated plug-and-play three-operator splitting methods for nonconvex optimization with applications to image restoration,'' \emph{SIAM Journal on Imaging Sciences}, vol.~17, no.~2, pp. 1145--1181, 2024.

\bibitem{deng2025deepsn}
X.~Deng, C.~Zhang, L.~Jiang, J.~Xia, and M.~Xu, ``Deep{SN}-{N}et: Deep semi-smooth {N}ewton driven network for blind image restoration,'' \emph{IEEE Transactions on Pattern Analysis and Machine Intelligence}, vol.~47, no.~4, pp. 2632--2646, 2025.

\bibitem{shlezinger2023model}
N.~Shlezinger, J.~Whang, Y.~C. Eldar, and A.~G. Dimakis, ``Model-based deep learning,'' \emph{Proceedings of the IEEE}, vol. 111, no.~5, pp. 465--499, 2023.

\bibitem{joukovsky2023interpretable}
B.~Joukovsky, Y.~C. Eldar, and N.~Deligiannis, ``Interpretable neural networks for video separation: Deep unfolding {RPCA} with foreground masking,'' \emph{IEEE Transactions on Image Processing}, vol.~33, pp. 108--122, 2023.

\bibitem{xiong2025drpca}
Z.~Xiong, F.~Zhou, F.~Wu, S.~Yuan, M.~Fu, Z.~Peng, J.~Yang, and Y.~Dai, ``{DRPCA}-{N}et: Make robust {PCA} great again for infrared small target detection,'' \emph{IEEE Transactions on Geoscience and Remote Sensing}, vol.~63, pp. 1--16, 2025.

\bibitem{candes2011robust}
E.~J. Cand{\`e}s, X.~Li, Y.~Ma, and J.~Wright, ``Robust principal component analysis?'' \emph{Journal of the ACM (JACM)}, vol.~58, no.~3, pp. 1--37, 2011.

\bibitem{wu2025rpcanet++}
F.~Wu, Y.~Dai, T.~Zhang, Y.~Ding, J.~Yang, M.-M. Cheng, and Z.~Peng, ``{RPCANet}++: Deep interpretable robust {PCA} for sparse object segmentation,'' \emph{arXiv preprint arXiv:2508.04190}, 2025.

\bibitem{elad2023image}
M.~Elad, B.~Kawar, and G.~Vaksman, ``Image denoising: The deep learning revolution and beyond—a survey paper,'' \emph{SIAM Journal on Imaging Sciences}, vol.~16, no.~3, pp. 1594--1654, 2023.

\bibitem{chen2024learning}
X.~Chen, J.~Liu, and W.~Yin, ``Learning to optimize: A tutorial for continuous and mixed-integer optimization,'' \emph{Science China Mathematics}, vol.~67, no.~6, pp. 1191--1262, 2024.

\bibitem{kong2021deep}
S.~Kong, W.~Wang, X.~Feng, and X.~Jia, ``Deep {RED} unfolding network for image restoration,'' \emph{IEEE Transactions on Image Processing}, vol.~31, pp. 852--867, 2021.

\bibitem{de2024deep}
B.~De~Weerdt, Y.~C. Eldar, and N.~Deligiannis, ``Deep unfolding transformers for sparse recovery of video,'' \emph{IEEE Transactions on Signal Processing}, vol.~72, pp. 1782--1796, 2024.

\bibitem{fu2024rotation}
J.~Fu, Q.~Xie, D.~Meng, and Z.~Xu, ``Rotation equivariant proximal operator for deep unfolding methods in image restoration,'' \emph{IEEE Transactions on Pattern Analysis and Machine Intelligence}, vol.~46, no.~10, pp. 6577--6593, 2024.

\bibitem{gregor2010learning}
K.~Gregor and Y.~LeCun, ``Learning fast approximations of sparse coding,'' in \emph{Proceedings of the 27th International Conference on International Conference on Machine Learning}, 2010, pp. 399--406.

\bibitem{yang2020admm}
Y.~Yang, J.~Sun, H.~Li, and Z.~Xu, ``{ADMM}-{CSNet}: A deep learning approach for image compressive sensing,'' \emph{IEEE Transactions on Pattern Analysis and Machine Intelligence}, vol.~42, no.~3, pp. 521--538, 2020.

\bibitem{zhang2018ista}
J.~Zhang and B.~Ghanem, ``{ISTA-N}et: Interpretable optimization-inspired deep network for image compressive sensing,'' in \emph{2018 IEEE/CVF Conference on Computer Vision and Pattern Recognition}, 2018, pp. 1828--1837.

\bibitem{you2021ista}
D.~You, J.~Xie, and J.~Zhang, ``{ISTA-N}et++: Flexible deep unfolding network for compressive sensing,'' in \emph{2021 IEEE International Conference on Multimedia and Expo (ICME)}.\hskip 1em plus 0.5em minus 0.4em\relax IEEE, 2021, pp. 1--6.

\bibitem{han2025dista}
S.~Han, S.~Yang, X.~Zhang, Y.~Li, X.~Li, J.~Yang, M.-M. Cheng, and Y.~Dai, ``{DISTA-N}et: Dynamic closely-spaced infrared small target unmixing,'' \emph{arXiv preprint arXiv:2505.19148}, 2025.

\bibitem{vaswani2017attention}
A.~Vaswani, N.~Shazeer, N.~Parmar, J.~Uszkoreit, L.~Jones, A.~N. Gomez, {\L}.~Kaiser, and I.~Polosukhin, ``Attention is all you need,'' \emph{Advances in Neural Information Processing Systems}, vol.~30, 2017.

\bibitem{guo2023beyond}
M.-H. Guo, Z.-N. Liu, T.-J. Mu, and S.-M. Hu, ``Beyond self-attention: External attention using two linear layers for visual tasks,'' \emph{IEEE Transactions on Pattern Analysis and Machine Intelligence}, vol.~45, no.~5, pp. 5436--5447, 2023.

\bibitem{huang2019ccnet}
Z.~Huang, X.~Wang, L.~Huang, C.~Huang, Y.~Wei, and W.~Liu, ``Ccnet: Criss-cross attention for semantic segmentation,'' in \emph{Proceedings of the IEEE/CVF International Conference on Computer Vision}, 2019, pp. 603--612.

\bibitem{woo2018cbam}
S.~Woo, J.~Park, J.-Y. Lee, and I.~S. Kweon, ``{CBAM}: Convolutional block attention module,'' in \emph{Proceedings of the European Conference on Computer Vision}, 2018, pp. 3--19.

\bibitem{yang2025mtmlnet}
B.~Yang, F.~Li, S.~Zhao, W.~Wang, J.~Luo, H.~Pu, M.~Zhou, and Y.~Pi, ``{MTMLN}et: Multi-task mutual learning network for infrared small target detection and segmentation,'' \emph{IEEE Transactions on Image Processing}, vol.~34, pp. 4414--4425, 2025.

\bibitem{zhang2024irsam}
M.~Zhang, Y.~Wang, J.~Guo, Y.~Li, X.~Gao, and J.~Zhang, ``{IRSAM}: Advancing segment anything model for infrared small target detection,'' in \emph{European Conference on Computer Vision}.\hskip 1em plus 0.5em minus 0.4em\relax Springer, 2024, pp. 233--249.

\bibitem{zhang2025saist}
M.~Zhang, X.~Li, F.~Gao, J.~Guo, X.~Gao, and J.~Zhang, ``{SAIST}: Segment any infrared small target model guided by contrastive language-image pretraining,'' in \emph{Proceedings of the Computer Vision and Pattern Recognition Conference}, 2025, pp. 9549--9558.

\bibitem{bouwmans2017decomposition}
T.~Bouwmans, A.~Sobral, S.~Javed, S.~K. Jung, and E.-H. Zahzah, ``Decomposition into low-rank plus additive matrices for background/foreground separation: A review for a comparative evaluation with a large-scale dataset,'' \emph{Computer Science Review}, vol.~23, pp. 1--71, 2017.

\bibitem{cai2010singular}
J.-F. Cai, E.~J. Cand{\`e}s, and Z.~Shen, ``A singular value thresholding algorithm for matrix completion,'' \emph{SIAM Journal on Optimization}, vol.~20, no.~4, pp. 1956--1982, 2010.

\bibitem{gu2017weighted}
S.~Gu, Q.~Xie, D.~Meng, W.~Zuo, X.~Feng, and L.~Zhang, ``Weighted nuclear norm minimization and its applications to low level vision,'' \emph{International Journal of Computer Vision}, vol. 121, no.~2, pp. 183--208, 2017.

\bibitem{yao2022low}
Q.~Yao, Y.~Wang, B.~Han, and J.~T. Kwok, ``Low-rank tensor learning with nonconvex overlapped nuclear norm regularization,'' \emph{Journal of Machine Learning Research}, vol.~23, no. 136, pp. 1--60, 2022.

\bibitem{nair2010rectified}
V.~Nair and G.~E. Hinton, ``Rectified linear units improve restricted boltzmann machines,'' in \emph{Proceedings of the 27th International Conference on Machine Learning (ICML-10)}, 2010, pp. 807--814.

\bibitem{beck2009fast}
A.~Beck and M.~Teboulle, ``A fast iterative shrinkage-thresholding algorithm for linear inverse problems,'' \emph{SIAM Journal on Imaging Sciences}, vol.~2, no.~1, pp. 183--202, 2009.

\bibitem{li2018highly}
X.~Li, D.~Sun, and K.-C. Toh, ``A highly efficient semismooth {N}ewton augmented {L}agrangian method for solving {L}asso problems,'' \emph{SIAM Journal on Optimization}, vol.~28, no.~1, pp. 433--458, 2018.

\bibitem{virmaux2018lipschitz}
A.~Virmaux and K.~Scaman, ``Lipschitz regularity of deep neural networks: analysis and efficient estimation,'' \emph{Advances in Neural Information Processing Systems}, vol.~31, 2018.

\bibitem{zhang2018ffdnet}
K.~Zhang, W.~Zuo, and L.~Zhang, ``{FFDN}et: Toward a fast and flexible solution for {CNN}-based image denoising,'' \emph{IEEE Transactions on Image Processing}, vol.~27, no.~9, pp. 4608--4622, 2018.

\bibitem{rahman2016optimizing}
M.~A. Rahman and Y.~Wang, ``Optimizing intersection-over-union in deep neural networks for image segmentation,'' in \emph{International Symposium on Visual Computing}.\hskip 1em plus 0.5em minus 0.4em\relax Springer, 2016, pp. 234--244.

\end{thebibliography}

\end{document}